\documentclass[letterpaper]{article} 
\usepackage{aaai2026}  
\nocopyright
\usepackage[hyphens]{url}  
\usepackage{graphicx} 
\usepackage{natbib}  
\usepackage{caption} 
\usepackage{algorithm}
\usepackage{amsmath}
\usepackage{dsfont}

\usepackage{microtype}
\usepackage{graphicx}
\usepackage{booktabs} 
\usepackage{multirow}
\usepackage[table,xcdraw]{xcolor}
\usepackage{arydshln}  
\usepackage{pifont}
\usepackage[table,xcdraw]{xcolor}

\definecolor{light-gray0}{gray}{0.9}
\usepackage{caption}
\usepackage{xcolor}

\usepackage{amsmath}
\usepackage{amssymb}
\usepackage{mathtools}
\usepackage{amsthm}
\usepackage{subcaption} 
\usepackage{arydshln}
\usepackage{algpseudocode}

\usepackage{newfloat}
\usepackage{listings}
\DeclareCaptionStyle{ruled}{labelfont=normalfont,labelsep=colon,strut=off} 
\floatstyle{ruled}
\newfloat{listing}{tb}{lst}{}
\floatname{listing}{Listing}

\usepackage{booktabs}
\title{Visual Distortion Detection in UGC Images Using Large Multimodal Models}
\author {
    Ziheng~Jia\*\textsuperscript{\rm 1}, 
    Yingji~Liang\*\textsuperscript{\rm 2},
    Jiaying~Qian\textsuperscript{\rm 1},
    Xiongkuo~Min\textsuperscript{\rm 1}\thanks{Corresponding author.}
}
\affiliations {
    \textsuperscript{\rm 1}Shanghai Jiao Tong University\\
    \textsuperscript{\rm 2}East China Normal University\\
    jzhws1@sjtu.edu.cn
}

\begin{document}

\maketitle

\begin{abstract}
The localized depiction of perceptual quality has long been a crucial, yet underexplored, challenge in image quality assessment (IQA). Existing approaches based on large multimodal models (LMMs) predominantly rely on text-driven supervised fine-tuning (SFT).
 However, this training paradigm exhibits notable limitations in detection accuracy. Moreover,  synthetically distorted images, which are often used as the primary training data source,
 show a significant generalization gap when deployed in real-world scenarios; thus, the \textbf{synthetic-to-authentic (\textit{S2A})} problem represents a critical challenge.
Motivated by these issues, we propose \textbf{\textit{VIGIL}}, which leverages the LMM architecture for precise visual distortion detection. From a candidate pool of over $1000K$ samples, we construct the \textbf{\textit{VIGIL-140K}} training set, which consists of over $140K$ distorted images. These images are obtained through rigorous quality filtering and carefully crafted distortion injection, covering $8$ major synthetic distortion categories.
 Our model leverages different layers of the large language model (LLM) decoder, treating them as \textit{multiple detectors} that perform synchronous distortion detection using multi-level features. Additionally, we retain distortion cues from predictions assigned to the non-distortion class, which helps mitigate the ambiguous foreground-background (\textit{FG-BG}) separation commonly encountered in the \textit{S2A} problem.
 After post-processing, our model consistently outperforms strong baselines on both in-domain synthetic distortion detection and \textit{S2A} tasks.
\end{abstract}


\section{Introduction}
Image quality assessment (IQA) is one of the most extensively studied topics in computer vision. Classic IQA research mostly concentrates on perceptual quality rating, where the model's output is aligned with the mean opinion score (MOS) derived from human subjective experiments~\cite{zhai2020perceptual}.
 With the recent proliferation of large multimodal models (LMMs), IQA research on user-generated content (UGC) images has increasingly leveraged the versatility of LMMs to enable comprehensive quality analysis~\cite{zhang2025large,AIBench}. However, most existing works still focus on \textbf{global} quality assessment, such as providing an overall quality score (first impression) or global descriptions aligned with specific quality factors and attributes. In contrast, for images with significant \textbf{localized} artifacts, systematic detection methods are still underdeveloped.
 This naturally leads to the question:
\begin{figure}[h]
    \centering
    \includegraphics[width=0.98\linewidth]{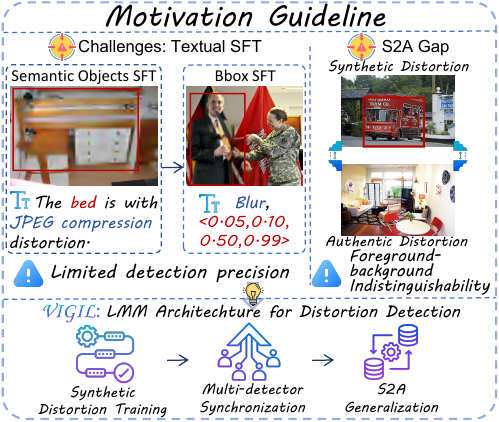}
    \caption{The limited detection precision caused by textual SFT and
  the \textit{FG-BG} indistinguishability in \textit{S2A} generalization  tasks are two major challenges for existing models.
To address these issues, we propose \textit{VIGIL}. It leverages synthetic distortion training for scalable data expansion, employs multiple detectors for synchronous detection, and retains background class prediction information to improve \textit{S2A} generalization.
} 
    \label{fig:motivation}
\end{figure}
\textit{Why is it necessary to perform localized distortion detection and analysis beyond global assessment for UGC images?}

First, in-the-wild UGC images are more likely to be affected by spatial local distortions caused by hardware limitations or specific capture conditions compared to other image types like AI-generated content (AIGC) or professionally generated content (PGC).
 Accurately identifying and localizing such distortions holds significant \textbf{practical value}.
 Secondly, with the continuous improvement in the  perceptual quality of UGC images in recent years, meeting ``high-quality'' requirements is not merely equivalent to having a favorable global subjective impression; the focus of enhancement and post-processing has increasingly shifted towards fine-grained local details. As a result, it is becoming essential to evolve from global assessments to localized distortion detection, thereby enabling more effective \textbf{feedback-driven optimization}.
Third, many UGC images deliberately introduce effects (e.g., defocus blur, color shifts) in specific regions to improve aesthetics. Detecting these localized effects facilitates in-depth annotation, which, in turn, enhances the LMM's overall \textbf{quality understanding}.

Some existing LMM-based approaches for distortion detection primarily rely on \textbf{text generation}~\cite{chen2024grounding}: they use supervised fine-tuning (SFT) with data labeled with regional distortion details, enabling LMMs to generate localized distortion cues.
 This training paradigm typically takes two forms: \textbf{semantic-object-level} descriptions, where the training label includes quality descriptions linked to local semantic objects with injected distortions~\cite{jia2025refine,jia2026scaling}, and \textbf{bounding-box (bbox)-level}~\cite{chen2024grounding}, where coordinates are used directly as the SFT labels.
 However, in real-world scenarios, the locally distorted regions are challenging to define using explicit semantic objects. Furthermore, directly generating distortion coordinates without a clear matching and classification process can significantly reduce \textbf{detection accuracy}.
 These factors collectively constrain the localized distortion detection capability in existing models.
 
Another significant challenge is \textbf{generalization in real-world scenarios}. While synthesizing distortions efficiently scales training datasets, the injected distortions, which typically follow regular shapes and uniform spatial intensity, differ significantly from authentic distortions in \textit{
foreground (FG)}-\textit{background (BG)} distinguishability.
 Consequently, the issue of generalizing from synthetic to authentic distortions (\textit{\textbf{S2A}}) requires careful consideration.

In response to the challenges, we propose \textbf{VIGIL}, which enables accurate detection of localized \textbf{\underline{vi}}sual distortions in U\textbf{\underline{G}}C \textbf{\underline{i}}mages using the \textbf{\underline{L}}MM architecture. Rather than relying on SFT, we revert to the classical object detection paradigm, leveraging LMM's powerful representation learning and regression capabilities.
 To accelerate convergence and improve the model's detection capability, we utilize multiple layers of the large language model (LLM) part for synchronous detection. To mitigate the \textit{\textit{FG-BG} indistinguishability} in \textit{S2A} tasks, we propose a simple yet effective technique that reuses detection cues from boxes predicted as non-distortion. The overview of our work is shown in Fig.~\ref{fig:motivation}.

Our main contributions are summarized as follows:
\begin{enumerate}
    \item We construct the \textbf{VIGIL-140K} dataset. From the source pool of over $1000K$ images, we filter those suitable for distortion injection, perform region-level distortion combination, and finally obtain a large-scale training set with over $200K$ distortion area labels.
    \item We introduce \textbf{VIGIL-8B}, an LMM-based model designed for perceiving and detecting local visual distortions. The model employs multiple detector synchronization, retains informative signals from boxes classified as background, and incorporates post-processing to refine the final outputs.
    \item Our model achieves state-of-the-art (SOTA) performance in both in-domain synthetic distortion and \textbf{out-of-domain (OOD) authentic distortion detection} tasks.
\end{enumerate}

\begin{figure*}
    \centering
    \includegraphics[width=0.945\linewidth]{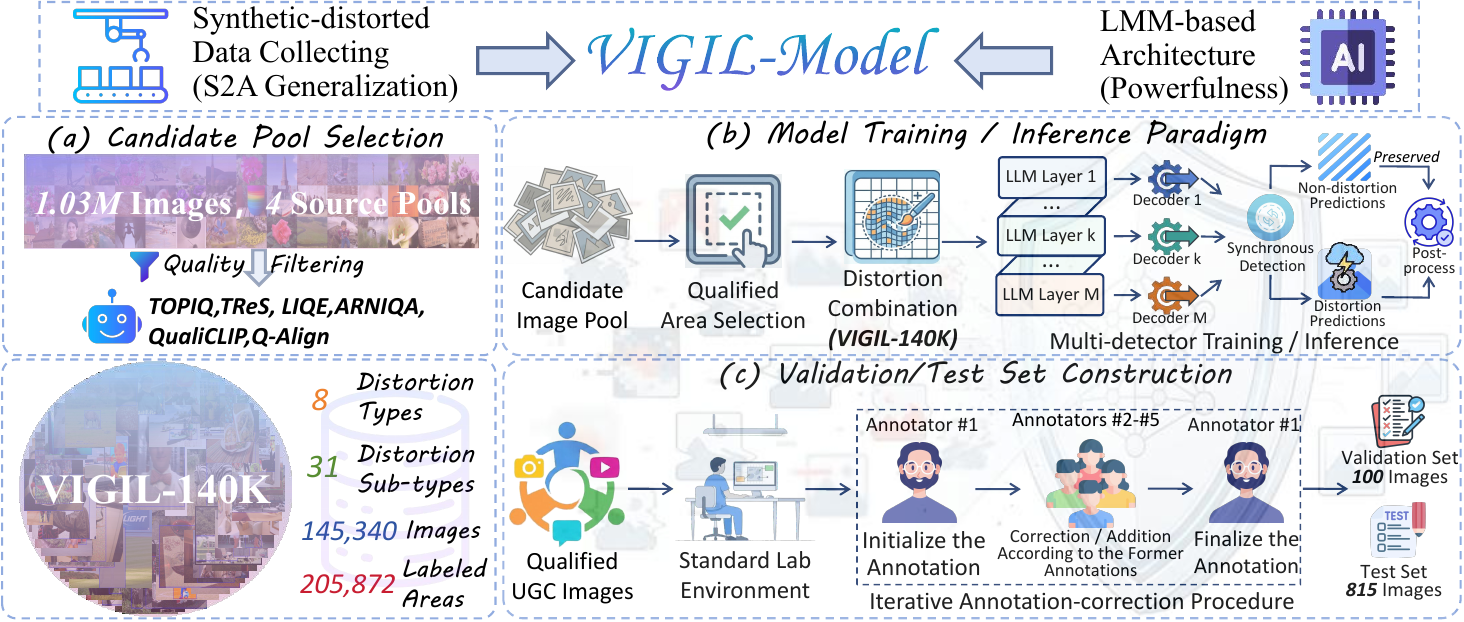}
    \caption{The main workflow of \textit{VIGIL}. } 
    \label{fig:workflow}
\end{figure*}
\begin{table}[!t]\tiny
    \centering
\renewcommand\arraystretch{1.2}
\renewcommand\tabcolsep{0.7pt}
\belowrulesep=0pt\aboverulesep=0pt
 
   \resizebox{0.97\linewidth}{!}{\begin{tabular}{c|c|c|c|c}
    \hline
    \textbf{Source} &\textbf{\# Images} &\textbf{\# AF } & \textbf{\# AD} & \textbf{\# Dist. Areas} \\ \hline
   \textit{\textbf{Online}}&$600K$&$28,369$ &$8,600$ &$10,353$ \\

   \textit{\textbf{CLIP-Pretrain}}&$300K$&$68,132$ &$54,940$&$75,559$ \\

   \textit{\textbf{COCO}}&$110K$&$95,768$ &$80,700$ &$117,706$ \\

    \textit{\textbf{Unsplash}}&$10K$&$9,783$ &$1,100$ &$2,254$\\
   
     \hline
     \textit{\textbf{Total}}&$1028K$&$202,052$ & $145,340$& $205,872$ \\
     \hline
\end{tabular}
 }
    \caption{Statistic summary of \textit{VIGIL-140K}. ``AF" means after the quality filtering. ``AD" indicates after the distortion combination process.}
\label{tab:dataStatistic}
\end{table}

\section{Related Works}
\subsection{Object Detection Models}

Object detection is a classic problem in computer vision.
\textit{Fast R-CNN}~\cite{girshick2015fast} computes convolutional features and applies \textit{RoI pooling} to obtain region features for classification and bbox regression. \textit{Faster R-CNN}~\cite{ren2016faster} introduces the \textit{region proposal network (RPN)}, which shares backbone features with the detection model, enabling two-stage detection without the need for external anchors. \textit{Detection Transformer (DETR)}~\cite{carion2020end} reformulates detection as a set prediction bipartite matching problem, removing the need for post-processing.
 \textit{Deformable DETR}~\cite{zhudeformable} adopts multi-scale \textit{deformable attention} that samples points around reference locations, accelerating convergence. \textit{YOLO-series}~\cite{redmon2016you} is a real-time object detection model series that processes the image in a single pass. \textit{DINO}~\cite{zhangdino}, and \textit{Grounding-DINO}~\cite{liu2024grounding} enhance \textit{DETR}-style detectors through improved query design and denoising-based training supervision.
\subsection{Image Local Distortion Detection}
\label{in-domain}
\textit{Q-Ground}~\cite{chen2024q} leverages LMMs with the grounding-training paradigm for image distortion segmentation. \textit{Grounding-IQA}~\cite{chen2024grounding} distills location-relevant information from existing human-annotated SFT data~\cite{wu2024q}. During detection, it adopts a training-inference scheme where the model directly outputs bbox coordinates. \textit{Refine-IQA-S1}~\cite{jia2025refine} employs the rule-based reinforcement learning strategy, using the \textit{Intersection over Union (IOU)} between the target boxes and the predicted boxes as the reward, while still relying on textual coordinate output during inference. \textit{ViDA-UGC}~\cite{liao2025vida} constructs a human-annotated dataset covering distortion detection, grounding, and description, and conducts \textit{chain-of-thought (CoT)}-based SFT.
However, these methods still exhibit limitations. First, they rely heavily on labor-intensive human annotation, with a relatively narrow range of distortion categories, and fail to fully achieve data scaling and explore the \textit{S2A} generalization. Second, the majority of these approaches depend on text generation, which limits both the accuracy and robustness of the localization process.

\section{The VIGIL}

In response to the above issues, the \textit{VIGIL} is trained entirely using synthetic distorted images. Its goal is to achieve high reliability in synthetic distortion detection while maximizing its \textit{\textit{S2A} generalization} capability in UGC scenarios. The overall workflow of  \textit{VIGIL} is demonstrated in Fig.~\ref{fig:workflow}.

\subsection{Training Data Preparation}
Synthetic distortions are applied to selected regions of high-quality images, maximizing diversity and randomness in both their locations and types.

\subsubsection{Candidate Image Pool Selection}
We select diverse UGC images as the source for our training dataset. Specifically, we choose \textit{COCO-2017-train}~\cite{lin2014microsoft}(containing $110K$ samples), a subset of \textit{CLIP}~\cite{radford2021learning} pretraining data (approximately $300K$ images), short video screenshots crawled from streaming platforms ($600K$ samples), and $10K$ high-quality aesthetic images from \textit{Unsplash}. These datasets span a wide range of semantic content, making them suitable for robust model training and \textit{S2A} generalization.

To prevent the influence of pre-existing distortions, we apply quality filtering.
 We use $6$ no-reference IQA models (\textit{TOPIQ-NR}~\cite{topiq}, \textit{TReS}~\cite{golestaneh2022no}, \textit{LIQE}~\cite{zhang2023blind}, \textit{ARNIQA}~\cite{agnolucci2024arniqa}, \textit{QualiCLIP}~\cite{agnolucci2024quality}, and \textit{Q-Align}~\cite{wu2024q1}) (pretrained on \textit{KonIQ-10K}~\cite{hosu2020koniq}) to score the entire source dataset. After normalizing them to a $[0,100)$ scale, we select data with scores above $85$. As these IQA methods for regular UGC images do not apply to the ultra-high-definition \textit{Unsplash} images, we manually check this part. We use a 4K resolution display device to present these images at their original resolution. Images with low quality or those that have undergone noticeable aesthetic post-processing (e.g., background defocusing) are excluded.

\subsubsection{Distortion Combination}
To ensure local distortions are perceptible, we further filter randomly picked local regions by \textbf{sharpness}, \textbf{colorfulness}, and \textbf{brightness}, retaining only those that satisfy the criteria (see supplementary materials (\textit{Supp.})). Following the commonly used \textit{KADIS-700K}~\cite{lin2019kadid}, we consider $10$ common distortion types: \textsc{``blur"}, \textsc{``noise"}, \textsc{``compression"}, \textsc{``overexposure"}, \textsc{``contrast strengthen"}, \textsc{``underexposure"}, \textsc{``contrast weaken"}, \textsc{``saturate strengthen"}, \textsc{``saturate weaken"}, and \textsc{``oversharpen"}. Based on perceptual similarity, we merge \textsc{``overexposure"} with \textsc{``contrast strengthen"} and \textsc{``underexposure"} with \textsc{``contrast weaken"}, yielding $8$ major categories (with $31$ sub-types detailed in \textit{Supp.}). For each category, we apply $2$ severity levels, \textsc{``noticeable"} and \textsc{``severe"}. Details of the distortion combination strategy are provided in Algorithm~\ref{distortion_combination} and \textit{Supp.}. For each image where local distortions are successfully added, we record the distortion category and the bbox of each modified region. The statistical information of the \textit{VIGIL-140K} is shown in Tab.~\ref{tab:dataStatistic} and also detailed in \textit{Supp.}.

\begin{algorithm}
\caption{Distortion Combination Process (per image)}
\begin{algorithmic}[1]
\State $K \gets \text{random integer between}  ~~1 ~~\text{and}~~ 3 $ 
\State $count \gets 0$
\While{$count < K$}
\State $rect\_area \gets \text{randomly select rectangle area between }$ 
\State $\frac{1}{20} \text{ and } 1 \text{ of the image}$
\State $distortion \gets \text{randomly select one distortion}$
\State $metrics \gets \text{calculate metrics for the selected region}$
\If{$rect\_area$: \textit{clarity}, \textit{brightness}, and \textit{colorfulness} $metrics$ are qualified}
\State \text{Add $distortion$, record \textit{type~/~bbox}}
\State $count \gets count + 1$
\Else \State \text{Skip this region}
\State $count \gets count + 1$
\EndIf
\EndWhile
\end{algorithmic}
\label{distortion_combination}
\end{algorithm}

\subsection{Model Structure Design}

LMMs typically exhibit strong regression capacity. Building on this property, we repurpose the LMM architecture as a powerful encoder to comprehensively capture low-level perceptual image features. 
We further exploit multiple layers from the LLM part as parallel detectors for synchronous detection. Finally, the detections for each image are produced after post-processing during the inference stage. The detailed model structure is shown in Fig.~\ref{fig:model}.

Following \textit{DETR}~\cite{carion2020end}, we first encode the image to obtain a sequence of image tokens, and then append \(N\) \textbf{learnable area query tokens}. In designing the attention mask, we apply causal attention to each image token while using global attention for all area query tokens.
 We choose \(M\) layers from the LLM part, each serving as an \textbf{independent} detector. We attach two linear heads for classification ($9$ classes, including the non-distortion class) and box regression after each selected layer, enabling up to \(M \times N\) predictions. As these detectors operate independently, we compute the matching cost for each detector’s predictions and apply \textit{Hungarian Matching}~\cite{kuhn1955hungarian} separately.
This design is motivated by two considerations. First, visual distortion detection largely relies on \textbf{low-level cues}, which makes representations from the early stages of the LLM decoder potentially more informative. Secondly, leveraging multiple detectors for synchronous detection can be viewed as integrating complementary \textbf{perceptual perspectives}. As a result, optimizing detection at earlier stages expands the pool of candidate predictions, potentially reducing missed detections. Compared to simply increasing the number of learnable area query  tokens, this approach also accelerates convergence.

Specifically, let \( y \) denote the ground-truth set of distorted areas and $\hat{y}_k=\left\{\hat{y}_{ki}\right\}_{i=1}^N$ the set of $N$ predictions of the \textit{$k$-th} detector. We treat \( y \) as a set of size \( N \), padded with \( \varnothing \) (representing the non-distortion class). For the \textit{$k$-th} detector, the \textit{Hungarian Matching} is represented as:
\begin{figure*}
    \centering
    \includegraphics[width=0.945\linewidth]{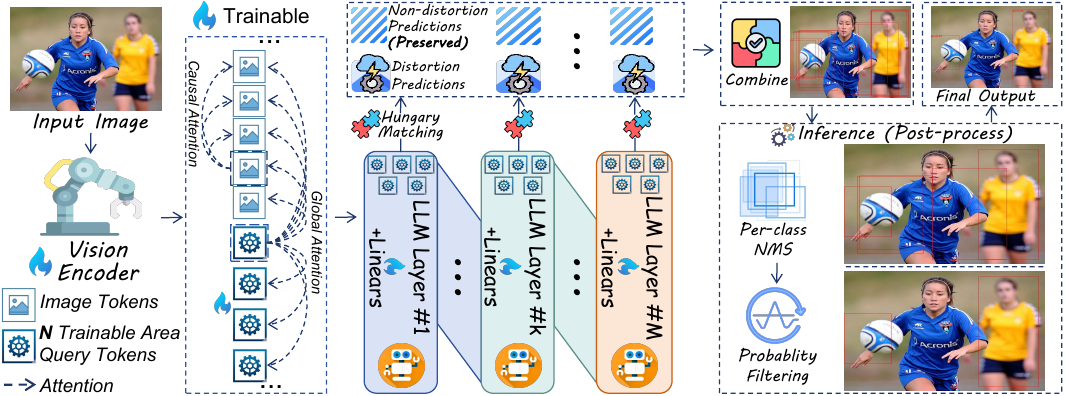}
    \caption{The model structure of \textit{VIGIL}.} 
    \label{fig:model}
\end{figure*}
\begin{equation}
    \hat{\sigma_k}=\underset{\sigma \in \mathfrak{S}_N}{\arg \min } \sum_{i=1}^N \mathcal{L}_{\text {match }}\left(y_i, \hat{y}_{k\sigma(i)}\right)\!,
\end{equation}
where $\mathcal{L}_{\text {match }}\left(y_i, \hat{y}_{k\sigma(i)}\right)$ is a pair-wise matching cost between ground truth $y_i$ and a prediction with its index (mapping) $\sigma(i)$ in the \textit{$k$-th} detector's predicted set. 
The matching cost accounts for both classification confidence and the alignment between the predicted bbox and the ground truth. Each ground-truth element \(i\) is represented as \(y_i=(c_i, b_i)\), where \(c_i\) denotes the target class label (possibly \(\varnothing\)) and \(b_i \in [0,1]^4\) specifies the ground-truth bbox in \textsc{``XYXY"} format, normalized by the image size. For the prediction indexed by \(\sigma_k(i)\), we denote the predicted probability of class \(c_i\) as \(\hat{p}_{\sigma_k(i)}(c_i)\) and the predicted box as \(\hat{b}_{k\sigma_k(i)}\). Under these definitions, the cost associated with matching \(y_i\) to \(\sigma_k(i)\) (which may not be the optimal one) is given by:
\begin{equation}
\begin{aligned}
\mathcal{L}_{\text{match}}\left(y_i,\hat{y}_{k\sigma_k(i)}\right)={}&-\mathds{1}_{\{c_i\neq\varnothing\}}\hat{p}_{k\sigma_k(i)}(c_i)\\
&+\mathds{1}_{\{c_i\neq\varnothing\}}\mathcal{L}_{\text{box}}\left(b_i,\hat{b}_{k\sigma_k(i)}\right).
\end{aligned}
\end{equation}
while $\mathcal{L}_{\text {box }}$ is denoted as:
\begin{equation}
   \mathcal{L}_{\text{box}}\!\!=\!\! \lambda_{\text {giou}} \mathcal{L}_{\text {giou }}\!\!\!\left(b_i, \hat{b}_{k\sigma_k(i)}\right)\!\!+\!\lambda_{\mathrm{L} 1}\!\left\|b_i-\hat{b}_{k\sigma_k(i)}\right\|_1\!.
\end{equation}
Here we set $\lambda_{\text {giou}}=2$ and $\lambda_{\mathrm{L} 1}=5$. The final detection loss is the average of all detector losses after bipartite matching :

\begin{equation}
\label{loss}
\begin{aligned}
\mathcal{L}(y,\hat{y})={}&\frac{1}{M}\frac{1}{N}\sum_{k=1}^{M}\sum_{i=1}^{N}\Bigl[-\log\hat{p}_{k\hat{\sigma}_k(i)}(c_i)\\
&+\mathds{1}_{\{c_i\neq\varnothing\}}\mathcal{L}_{\text{box}}\left(b_i,\hat{b}_{k\hat{\sigma}_k(i)}\right)\Bigr].
\end{aligned}
\end{equation}
In practice, when \(c_i = \varnothing\), we down-weight the corresponding log-probability term by a factor of \(0.05\) for data balancing.

\subsection{\textit{S2A} Generalization Trick and Post-processing}
Compared with synthetic distortions, manually localizing visual distortions in real-world UGC images is substantially more costly and less standardized. In particular, \textbf{local distortion regions are often irregular and non-uniform in shape and intensity}, which blurs the boundary between \textit{FG} and \textit{BG}. Under such conditions, detection models tend to incur elevated false negatives.
To alleviate this issue, we apply an \textit{S2A} generalization trick,  avoiding discarding predictions assigned to the non-distorted class. Instead, we take the distortion category with the \textbf{second-highest classification probability} and its associated bbox as a valid predicted candidate. This strategy enlarges the candidate set and reduces errors caused by ambiguous \textit{FG-BG} separation.

In our model, effective detection further relies on post-processing. For all candidate boxes produced, we assume equal opinion weights; this allows \textbf{probabilities from different detectors to be compared directly}. We first apply class-wise \textit{non-maximum suppression (NMS)} with the \textit{IoU} threshold \(T_{\mathrm{NMS}}\). We then discard candidates with probability below \(T_{\mathrm{Prob}}\). The remaining boxes are taken as the outputs.

\subsection{Validation~/~Test Set Construction}
\label{valtest}
We evaluate the model on two primary tasks: the in-domain synthetic distortion detection task and, \textbf{more critically}, the OOD generalization task on authentic distortion localization.

For in-domain evaluation, we generate \(3000\) synthetically distorted images from the undistorted images in \textit{KADIS-700K}. We further include \(1600\) images from \textit{COCO-2017-test} and \textit{Unsplash} that satisfy the quality criteria, using the same distortion synthesis pipeline.

For the OOD task, we pick images from \textit{COCO-2017-val} and \textit{COCO-2017-test}. Since consistent annotation of authentic distortions among multiple annotators is challenging without reference opinions, we implement an \textbf{iterative annotation-correction procedure}:
The first (leader) annotator marks perceptually obvious distortion regions in the image, each belonging to one of the $8$ major categories.
Annotators \textit{2-5} then review the marked regions in turn; they may \textit{revise} or \textit{remove} regions that are not perceptually valid and \textit{add regions} they consider missed. Finally, the leader annotator performs a last-pass check, during which only deletions are allowed.

All annotations are produced using \textit{LabelMe}. A region is annotated only if the distortion is perceptually salient, and the four-connected distortion area is expected to align reasonably with the annotated bbox. Each image contains \(1\!-\!3\) distorted regions, allowing overlaps. The final dataset consists of $915$ human-labeled UGC images. We reserve \(100\) images for validation, and use the remaining $815$ for testing.

\subsection{Training Details}
We use \textit{InternVL-3-8B-Instruct}~\cite{zhu2025internvl3} (LLM: \textit{Qwen2.5-7B}~\cite{team2024qwen2}) (with added learnable area query tokens and the linear classifiers and regressors) as the base model. The number of epochs is set to $1$, with checkpoint saving every $500$ step. Each saved checkpoint is evaluated on the validation set, and the \textit{AP50} metric of all labeled regions in the validation set is recorded. The checkpoint with the highest \textit{AP50} is chosen as the final model for the selected hyperparameter setting. More detailed model structure and hyperparameter settings are presented in the \textit{Supp.}.

\section{Experiments}
To thoroughly evaluate the performance of our model, we conduct comparative experiments against multiple strong baselines on both in-domain synthetic distortion detection and OOD authentic distortion detection tasks. In addition, to analyze the impact of key attributes on model performance, we perform comprehensive ablation studies accompanied by detailed discussions and analyses.
\begin{table*}[t]\small
    \centering
    \renewcommand\arraystretch{1.1}
    \setlength{\tabcolsep}{0.2pt}
    \belowrulesep=0pt\aboverulesep=0pt

    \resizebox{\linewidth}{!}{%
    \begin{tabular}{l|cccccccccccccccc|c|c}
    \hline
    \multicolumn{1}{l|}{\textbf{Dist. Type}}
      & \multicolumn{2}{c}{\textbf{Blur}}
      & \multicolumn{2}{c}{\textbf{Noise}}
      & \multicolumn{2}{c}{\textbf{Comp.}}
      & \multicolumn{2}{c}{\textbf{OE}}
      & \multicolumn{2}{c}{\textbf{UE}}
      & \multicolumn{2}{c}{\textbf{OS}}
      & \multicolumn{2}{c}{\textbf{US}}  & \multicolumn{2}{c|}{\textbf{OSH}}&\multirow{3}{*}{\textit{\textbf{mAP$\uparrow$}}}&\multirow{3}{*}{\textit{\textbf{AP50$\uparrow$}}}\\
    \cline{1-17}
    \multicolumn{1}{l|}{\textbf{\# of GT Areas}}
      & \multicolumn{2}{c}{\textbf{1,619}}
      & \multicolumn{2}{c}{\textbf{1,768}}
      & \multicolumn{2}{c}{\textbf{1,067}}
      & \multicolumn{2}{c}{\textbf{1,640}}
      & \multicolumn{2}{c}{\textbf{1,647}}
      & \multicolumn{2}{c}{\textbf{289}}
      & \multicolumn{2}{c}{\textbf{315}}  & \multicolumn{2}{c|}{\textbf{818}}&&\\
    \cline{1-17}
      \textbf{Models} & \textit{AP50}$\uparrow$ & \textit{AP75}$\uparrow$
      & \textit{AP50}$\uparrow$ & \textit{AP75}$\uparrow$
      & \textit{AP50}$\uparrow$ & \textit{AP75}$\uparrow$
      & \textit{AP50}$\uparrow$ & \textit{AP75}$\uparrow$
      & \textit{AP50}$\uparrow$ & \textit{AP75}$\uparrow$
      & \textit{AP50}$\uparrow$ & \textit{AP75}$\uparrow$ &\textit{AP50}$\uparrow$ & \textit{AP75}$\uparrow$&\textit{AP50}$\uparrow$ & \textit{AP75}$\uparrow$ &&\\
    \cdashline{1-19}
      \multicolumn{1}{l}{\textit{General LMMs}}\\ 
      \cdashline{1-19}
      \textsc{InternVL3.5-8B} &0.316&0.194&0.336&0.228&0.332&0.190&0.323&0.206&0.344&0.249&0.372&0.230&0.342&0.236&0.329&0.206&0.274&0.326\\ 
        \rowcolor{light-gray0}  \textsc{InternVL3.5-14B}&0.381&0.220&0.412&0.263&0.364&0.224&0.400&0.256&0.419&0.281&0.366&0.284&0.484&0.188&0.393&0.249&0.322&0.389\\ 
      \textsc{InternVL3.5-38B} &0.461&0.237&0.462&0.315&0.469&0.254&0.448&0.270&0.478&0.335&0.520&0.269&0.520&0.312&0.475&0.292&0.369&0.461\\ 
         \rowcolor{light-gray0} \textsc{Qwen3vl-8B}&0.298&0.171&0.319&0.224&0.316&0.177&0.304&0.194&0.326&0.228&0.352&0.189&0.340&0.189&0.329&0.178&0.259&0.313\\ 
           \textsc{Qwen3vl-32B}&0.448&0.256&0.463&0.291&0.449&0.243&0.411&0.264&0.447&0.321&0.457&0.293&0.462&0.293&0.430&0.253&0.350&0.437\\ 
           \rowcolor{light-gray0} \textsc{LLaVA-OV-1.5-7B}&0.301&0.159&0.305&0.203&0.306&0.191&0.266&0.186&0.320&0.231&0.318&0.213&0.304&0.170&0.319&0.195&0.249&0.297\\
           \textsc{GPT-4o (24-11-20)}&0.340&0.174&0.359&0.220&0.319&0.207&0.346&0.231&0.363&0.248&0.358&0.216&0.314&0.222&0.325&0.231&0.284&0.340\\
           \rowcolor{light-gray0} \textsc{GPT-5 (25-08-07)}&0.312&0.170&0.313&0.197&0.309&0.171&0.306&0.217&0.316&0.214&0.309&0.228&0.299&0.204&0.336&0.196&0.258&0.308\\
           \textsc{Gemini-3.0-Pro}&0.449&0.231&0.447&0.265&0.489&0.259&0.423&0.276&0.426&0.335&0.481&0.334&0.478&0.318&0.435&0.272&0.360&0.438\\
      \cdashline{1-19}
      \multicolumn{1}{l}{\textit{Detection Models}}\\ 
      \cdashline{1-19}
       \rowcolor{light-gray0} \textsc{Faster-RCNN-R50}&0.532&0.297&0.562&0.342&0.510&0.304&0.539&0.318&0.541&0.398&0.553&0.305&0.534&0.354&0.502&0.292&0.418&0.528  \\ 
        \textsc{DETR-R50}&0.706&0.390&0.747&0.484&0.734&0.397&0.692&0.479&0.737&0.511&0.748&0.471&0.731&0.459&0.723&0.456&0.601&0.719 \\ 
         \rowcolor{light-gray0} \textsc{Deformable-DETR-R50}&0.675&0.383&0.717&0.472&0.715&0.421&0.693&0.439&0.721&0.519&0.685&0.489&0.680&0.426&0.671&0.428&0.574&0.690 \\ 
         \textsc{YOLO-V11} &\textbf{0.865}&0.486&\underline{0.902}&\underline{0.581}&\textbf{0.882}&\underline{0.502}&\textbf{0.867}&\underline{0.546}&\underline{0.906}&\underline{0.628}&\underline{0.882}&0.553&\underline{0.909}&\underline{0.560}&\underline{0.853}&\underline{0.526}&\underline{0.708}&\underline{0.878} \\ 
       \rowcolor{light-gray0} \textsc{Grounding-DINO-R50} &0.829&\underline{0.488}&0.874&0.558&0.855&0.492&0.848&0.535&0.855&0.603&0.854&\underline{0.554}&0.881&0.558&0.812&0.512&0.687&0.851  \\ 
      \cdashline{1-19}
      
      \multicolumn{1}{l}{\textit{In-domain LMMs}}\\ 
      \cdashline{1-19}
    \textsc{Refine-IQA-S1} &0.636&0.374&0.694&0.443&0.634&0.377&0.635&0.420&0.688&0.442&0.651&0.424&0.733&0.459&0.626&0.385&0.540&0.652
     \\
    \rowcolor{light-gray0} \textbf{\textsc{VIGIL-8B}} &\underline{0.845}&\textbf{0.501}&\textbf{0.910}&\textbf{0.603}&\underline{0.869}&\textbf{0.524}&\underline{0.866}&\textbf{0.578}&\textbf{0.915}&\textbf{0.657}&\textbf{0.888}&\textbf{0.583}&\textbf{0.931}&\textbf{0.597}&\textbf{0.868}&\textbf{0.555}&\textbf{0.736}&\textbf{0.879}
      \\
     
    \cline{1-19}
    \end{tabular}%
    }
     \caption{Performance on the \textit{synthetic distortion detection} task. \textit{``Comp."}, \textit{``OE"}, \textit{``UE"}, \textit{``OS"}, \textit{``US"}, and \textit{``OSH"} represent \textit{``Compression"}, \textit{``Overexposure"}, \textit{``Underexposure"}, \textit{``Oversaturate"},\textit{``Undersaturate"}, and \textit{``Oversharpen"} respectively. Detection models have been trained on  \textit{VIGIL-140K}. ``\textit{R50}" denotes that the backbone of this model is \textit{ResNet-50}. [Per column: highest values are in \textbf{bold}, and second-highest values are \underline{underlined}.] }
    \label{tab:synthetic}
\end{table*}

\begin{table*}[!t]\small
    \centering
    \renewcommand\arraystretch{1.1}
    \setlength{\tabcolsep}{0.2pt}
    \belowrulesep=0pt\aboverulesep=0pt

    \resizebox{\linewidth}{!}{%
    \begin{tabular}{l|cccccccccccccccc|c|c}
    \hline
    \multicolumn{1}{l|}{\textbf{Dist. Type}}
      & \multicolumn{2}{c}{\textbf{Blur}}
      & \multicolumn{2}{c}{\textbf{Noise}}
      & \multicolumn{2}{c}{\textbf{Comp.}}
      & \multicolumn{2}{c}{\textbf{OE}}
      & \multicolumn{2}{c}{\textbf{UE}}
      & \multicolumn{2}{c}{\textbf{OS}}
      & \multicolumn{2}{c}{\textbf{US}}  & \multicolumn{2}{c|}{\textbf{OSH}}&\multirow{3}{*}{\textit{\textbf{mAP$\uparrow$}}}&\multirow{3}{*}{\textit{\textbf{AP50$\uparrow$}}}\\
    \cline{1-17}
    \multicolumn{1}{l|}{\textbf{\# of GT Areas}}
      & \multicolumn{2}{c}{\textbf{464}}
      & \multicolumn{2}{c}{\textbf{69}}
      & \multicolumn{2}{c}{\textbf{52}}
      & \multicolumn{2}{c}{\textbf{337}}
      & \multicolumn{2}{c}{\textbf{168}}
      & \multicolumn{2}{c}{\textbf{47}}
      & \multicolumn{2}{c}{\textbf{93}}  & \multicolumn{2}{c|}{\textbf{51}}&&\\
    \cline{1-17}
      \textbf{Models} & \textit{AP50}$\uparrow$ & \textit{AP75}$\uparrow$
      & \textit{AP50}$\uparrow$ & \textit{AP75}$\uparrow$
      & \textit{AP50}$\uparrow$ & \textit{AP75}$\uparrow$
      & \textit{AP50}$\uparrow$ & \textit{AP75}$\uparrow$
      & \textit{AP50}$\uparrow$ & \textit{AP75}$\uparrow$
      & \textit{AP50}$\uparrow$ & \textit{AP75}$\uparrow$ &\textit{AP50}$\uparrow$ & \textit{AP75}$\uparrow$&\textit{AP50}$\uparrow$ & \textit{AP75}$\uparrow$ &&\\
    \cdashline{1-19}
      \multicolumn{1}{l}{\textit{General LMMs}}\\ 
      \cdashline{1-19}
      \textsc{InternVL3.5-8B}&0.188&0.136&0.111&0.095&0.047&0.033&0.087&0.104&0.080&0.055&0.242&0.101&0.207&0.179&0.135&0.155&0.137&0.128   \\ 
       \rowcolor{light-gray0} \textsc{InternVL3.5-14B}&0.223&0.181&0.124&0.112&0.099&0.108&0.209&0.107&0.062&0.066&0.267&0.253&0.256&0.220&0.232&0.291&0.184&0.182\\ 
        \textsc{InternVL3.5-38B}&0.191&0.170&0.133&0.053&0.064&0.084&0.146&0.128&0.115&0.051&0.377&0.147&0.158&0.249&0.165&0.226&0.169&0.153\\ 
         \rowcolor{light-gray0} \textsc{Qwen3vl-8B}&0.153&0.134&0.095&0.046&0.075&0.048&0.110&0.098&0.067&0.079&0.090&0.115&0.154&0.184&0.197&0.125&0.118&0.115\\ 
           \textsc{Qwen3vl-32B}&0.156&0.121&0.067&0.064&0.019&0.124&0.125&0.095&0.067&0.073&0.172&0.135&0.293&0.197&0.329&0.142&0.154&0.131\\ 
           \rowcolor{light-gray0} \textsc{LLaVA-OV-1.5-7B}&0.143&0.138&0.111&0.140&0.109&0.056&0.139&0.125&0.061&0.064&0.199&0.135&0.277&0.235&0.128&0.209&0.146&0.129\\
            \textsc{GPT-4o (24-11-20)}&0.174&0.136&0.074&0.121&0.100&0.072&0.133&0.095&0.101&0.049&0.190&0.128&0.220&0.174&0.199&0.135&0.149&0.141\\
           \rowcolor{light-gray0} \textsc{GPT-5 (25-08-07)}&0.150&0.104&0.068&0.068&0.032&0.044&0.118&0.068&0.034&0.050&0.157&0.124&0.192&0.090&0.126&0.106&0.110&0.110\\
           \textsc{Gemini-3.0-Pro}&0.262&0.279&0.160&0.183&0.168&0.143&0.190&0.187&0.150&0.133&0.299&0.218&0.408&0.352&0.268&0.239&0.238&0.220\\
      \cdashline{1-19}
      \multicolumn{1}{l}{\textit{Detection Models}}\\ 
       \cdashline{1-19}
       \rowcolor{light-gray0} \textsc{Faster-RCNN-R50}&0.230&0.191&0.160&0.124&0.142&0.076&0.179&0.188&0.124&0.102&0.214&0.233&0.338&0.280&0.190&0.287&0.197&0.192  \\ 
\textsc{DETR-R50}&0.364&0.360&0.278&0.241&0.161&0.127&0.242&0.219&0.179&0.171&0.429&0.219&0.464&0.477&0.364&0.256&0.310&0.293  \\ 
         \rowcolor{light-gray0} \textsc{Deformable-DETR-R50}&0.416&0.329&0.218&0.232&0.190&0.132&0.237&0.235&0.153&0.139&0.330&0.322&0.497&0.461&0.360&0.458&0.300&0.302\\ 
         \textsc{YOLO-V11}&\underline{0.474}&0.389&0.305&0.230&0.177&\underline{0.178}&\underline{0.328}&0.271&0.197&0.180&0.434&\underline{0.418}&\underline{0.540}&\underline{0.602}&0.334&0.449&0.349&\underline{0.366}\\ 
            \rowcolor{light-gray0} \textsc{Grounding-DINO-R50} &0.469&\underline{0.428}&\underline{0.335}&\underline{0.287}&\underline{0.215}&0.140&0.310&\underline{0.311}&0.172&0.188&\underline{0.450}&0.320&0.538&0.549&\underline{0.411}&\underline{0.466}&\underline{0.362}&0.362  \\ 
      \cdashline{1-19}
      
      \multicolumn{1}{l}{\textit{In-domain LMMs}}\\ 
      \cdashline{1-19}
    \textsc{Refine-IQA-S1}  &0.321&0.270&0.131&0.117&0.113&0.115&0.231&0.200&0.133&0.131&0.389&0.258&0.452&0.326&0.260&0.317&0.254&0.250
      \\ 
    \rowcolor{light-gray0} \textbf{\textsc{VIGIL-8B}}&\textbf{0.617}&\textbf{0.545}&\textbf{0.420}&\textbf{0.323}&\textbf{0.237}&\textbf{0.228}&\textbf{0.444}&\textbf{0.374}&\textbf{0.275}&\textbf{0.241}&\textbf{0.596}&\textbf{0.483}&\textbf{0.800}&\textbf{0.742}&\textbf{0.597}&\textbf{0.594}&\textbf{0.482}&\textbf{0.502}
      \\
    \cline{1-19}
    \end{tabular}%
    }
     \caption{Performance on the \textit{authentic distortion detection} task. }
    \label{tab:authentic}
\end{table*}

\subsection{Experiments Settings and Main Results}
Both synthetic and authentic distortion detection are evaluated on the respective test sets mentioned in \textit{Validation~/~Test Set Construction} above. After a rigourous ablation study, we set $N=3$ and $M=5$ (using the \textit{$10$th}, \textit{$15$th}, \textit{$20$th}, \textit{$25$th}, and the final (\textit{$28$th}) layers of the LLM).  In the post-processing stage, we set \(T_{\mathrm{NMS}} = 0.3\), \(T_{\mathrm{Prob}} = 0.1\) for authentic distortion detection, and \(T_{\mathrm{NMS}} = 0.7\),\(T_{\mathrm{Prob}} = 0.8\) for synthetic distortion localization. In both experiments, we compare our method against comprehensive strong baselines. Since many general LMMs already support direct region localization, we include representative models from the newest \textit{Qwen3-VL}~\cite{li2026qwen3} series,  \textit{InternVL-3.5}~\cite{wang2025internvl3} series, and \textit{LLaVA-Onevision-1.5}~\cite{an2025llava}, as well as the proprietary \textit{GPT}~\cite{openai2024gpt4o,openai2025gpt5} and \textit{Gemini}~\cite{mallick2025gemini}. For these models, we formulate the input prompt as follows: 
\textit{``Please specify the exact area of distortions in the image using coordinates in the \textsc{`XYXY'} format (the top-left and bottom-right corners), normalized by the image height and width ($[0,1)$). The distortion types include: \textit{[eight major distortion types]}. The output format should be \textit{[category\_coordinates~/~category\_coordinates/...]}, and you may only provide up to $1$ to $3$ predicted distortion areas."
}. During evaluation, we directly select all predicted types and  regions (without post-processing) after the greedy search text generation process for reproduction.

Additionally, we compare our method against object detection models. These models are also trained on \textit{VIGIL-140K} and evaluated using their default inference protocols.
For in-domain distortion detection baselines, since most of these methods have certain limitations in terms of \textbf{task types} and \textbf{open-source availability}, which prevent a fair comparison with our model (we provide justification in \textit{Supp.}), we include only \textit{Refine-IQA-S1} for comparison.  We report \textit{AP50} and \textit{AP75} for each major distortion type, along with their \textit{mAP}. We also provide \textit{AP50} and \textit{AP75} on all the labeled areas.

From the results in Tabs.~\ref{tab:synthetic} (synthetic) and \ref{tab:authentic} (authentic), it is evident that general LMMs, which have not been specifically trained, perform poorly in both synthetic and \textit{S2A} tasks, especially in the latter. This indicates that text-generation-based methods may lack the accuracy and generalizability required for pixel-level perception-based localization tasks. 
When compared to other object detection models, \textbf{VIGIL-8B} demonstrates a clear performance advantage, particularly in the \textit{S2A} task. This underscores the strengths of LMM-based detection in terms of accuracy and reliability, as well as the improved \textit{S2A} generalization capability achieved through the application of \textit{S2A} generalization techniques.

\subsection{Discussions}
\label{discussion}
\subsubsection{Ablation Study}
First, we compare our training setup with the text-generation setup. For the latter, we rephrase the distortion regions in each training sample as a sequence of ``\textsc{distortion type} / $x_1, y_1, x_2, y_2$" format and implement SFT.  We record the \textit{mAP}, \textit{AP50}, and \textit{AP75} metrics on both synthetic and authentic tasks. Results are shown in Tab.~\ref{tab:setups}. 
The performance of the text-generation-based approach is significantly inferior to our settings on both tasks. We propose several reasons: first, the lack of effective matching processes in text generation, and second, training the box regression task as a token classification problem, which compromises detection accuracy. Moreover, the number of detectable areas in the text generation approach is constrained by the training data, limiting the model's generalization capability for detecting multiple distortion regions.

\begin{figure}[t]
    \centering
    \includegraphics[width=0.98\linewidth]{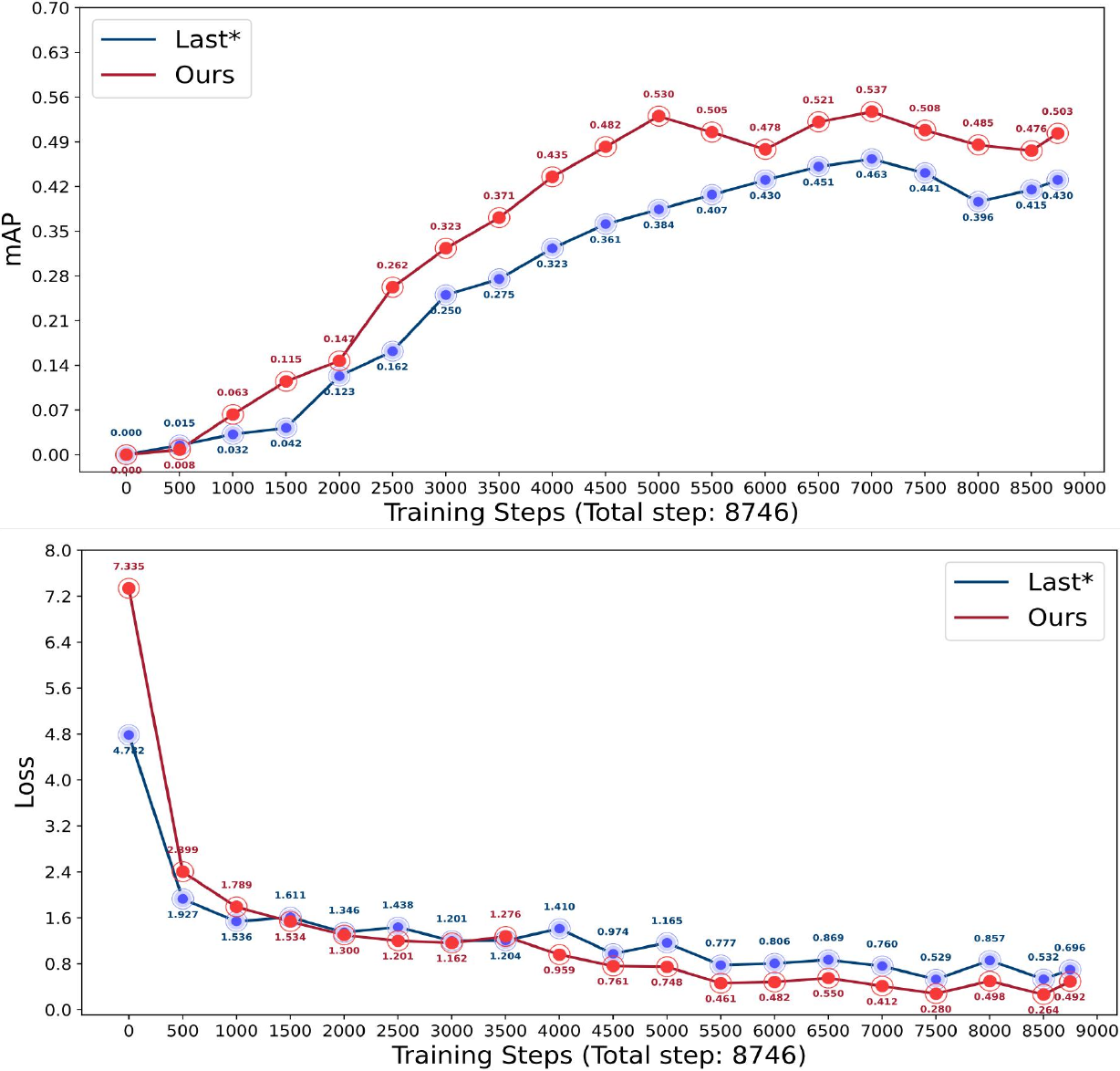}
    \caption{Curves of the variation of validation \textit{mAP} and training loss over the course of training.
} 
    \label{fig:curve}
\end{figure}

Next, we perform the ablation on the multi-detector setups. First, we retain all other training settings unchanged and utilize only the last LLM layer, referring to this model version as ``\textit{Last}". Additionally, we expand \(N = 15\) (equal to the areas predictable in our model) while still using only the last LLM layer, which we refer to as ``\textit{Last*}". We also record the loss convergence and the \textit{mAP} on the validation set during training for our setting and the ``\textit{Last*}" setting (shown in Fig.~\ref{fig:curve}). Furthermore, we retain the multi-detector setup, but unify the regression and classification linear modules attached to each detector with identical parameters, referred to as ``\textit{Unified}". Results are recorded in Tab.~\ref{tab:detector}. The multi-detector setup significantly improves the detection accuracy of the \textit{S2A} task compared to using only the final layer. Additionally, it converges faster during training and performs better than simply increasing the number of query tokens. Furthermore, using different decoder layers for each detector, paired with independent classification and regression heads, also positively impacts detection performance.

\begin{table}[!t]\small
    \centering
    \renewcommand\arraystretch{1.05}
    \setlength{\tabcolsep}{5pt}
    \belowrulesep=0pt\aboverulesep=0pt

    \resizebox{\linewidth}{!}{%
    \begin{tabular}{l|ccc|ccc}
    \hline
    \multirow{2}{*}{\textbf{Version}}
      & \multicolumn{3}{c|}{\textbf{Synthetic}}
      & \multicolumn{3}{c}{\textbf{Authentic}}\\

    \cline{2-7}
      &\textit{mAP}$\uparrow$ & \textit{AP50}$\uparrow$ & \textit{AP75}$\uparrow$ &
      \textit{mAP}$\uparrow$ & \textit{AP50}$\uparrow$ & \textit{AP75}$\uparrow$ \\
    \cline{1-7}
    \rowcolor{light-gray0}  \textit{Ours}&\textbf{0.736} &\textbf{0.879} &\textbf{0.574} &\textbf{0.482} &\textbf{0.502} &\textbf{0.439}\\
      \textit{Text}&0.457 &0.432 &0.263 &0.231 &0.225 &0.102\\
    \cline{1-7}
    \end{tabular}%
    }
    \caption{Training strategies ablation. [Per column: highest values are in \textbf{bold}.]}
    \label{tab:setups}
\end{table}

\begin{table}[!t]\small
    \centering
    \renewcommand\arraystretch{1.05}
    \setlength{\tabcolsep}{5pt}
    \belowrulesep=0pt\aboverulesep=0pt

    \resizebox{\linewidth}{!}{%
    \begin{tabular}{l|ccc|ccc}
    \hline
    \multirow{2}{*}{\textbf{Version}}
      & \multicolumn{3}{c|}{\textbf{Synthetic}}
      & \multicolumn{3}{c}{\textbf{Authentic}}\\

    \cline{2-7}
      &\textit{mAP}$\uparrow$ & \textit{AP50}$\uparrow$ & \textit{AP75}$\uparrow$ &
      \textit{mAP}$\uparrow$ & \textit{AP50}$\uparrow$ & \textit{AP75}$\uparrow$ \\
    \cline{1-7}
    \rowcolor{light-gray0}  \textit{Ours}&0.736 &0.879 &0.574 &\textbf{0.482} &\textbf{0.502} &\textbf{0.439} \\
      \textit{Last}&\textbf{0.757} &\textbf{0.913} &\textbf{0.582} &0.397 &0.408 &0.303 \\
    \rowcolor{light-gray0}  \textit{Last*}&0.743 &0.885 &0.569 &0.413 &0.420 &0.365 \\
      \textit{Unified}&0.735 &0.871 &0.562 &0.440 &0.451 &0.373\\
    \cline{1-7}
    \end{tabular}%
    }
     \caption{Ablation study on different multi-detector settings.}
    \label{tab:detector}
\end{table}

\begin{table}[!t]\small
    \centering
    \renewcommand\arraystretch{1.05}
    \setlength{\tabcolsep}{4.2pt}
    \belowrulesep=0pt\aboverulesep=0pt

    \resizebox{\linewidth}{!}{%
    \begin{tabular}{l|ccc|ccc}
    \hline
    \multirow{2}{*}{\textbf{Version}}
      & \multicolumn{3}{c|}{\textbf{Synthetic}}
      & \multicolumn{3}{c}{\textbf{Authentic}}\\

    \cline{2-7}
      &\textit{mAP}$\uparrow$ & \textit{AP50}$\uparrow$ & \textit{AP75}$\uparrow$ &
      \textit{mAP}$\uparrow$ & \textit{AP50}$\uparrow$ & \textit{AP75}$\uparrow$ \\
    \cline{1-7}
    \rowcolor{light-gray0}  \textit{Ours}&\textbf{0.736} &0.879 &\textbf{0.574} &\textbf{0.482} &\textbf{0.502} &\textbf{0.439}\\
      \textit{w~/~o \textit{S2A}} &0.727 &\textbf{0.882} &0.571 &0.407 &0.416 &0.351 \\
    \cline{1-7}
    \end{tabular}%
    }
      \caption{Ablation study on \textit{S2A} generalization tricks.}
    \label{tab:S2A}
\end{table}
\begin{table}[h]\small
    \centering
    \renewcommand\arraystretch{1.15}
    \setlength{\tabcolsep}{4.2pt}
    \belowrulesep=0pt\aboverulesep=0pt

    \resizebox{\linewidth}{!}{%
    \begin{tabular}{c|ccc|ccc}
    \hline
    \multirow{2}{*}{\textbf{Setting}}
      & \multicolumn{3}{c|}{\textbf{Synthetic}}
      & \multicolumn{3}{c}{\textbf{Authentic}}\\

    \cline{2-7}
      &\textit{mAP}$\uparrow$ & \textit{AP50}$\uparrow$ & \textit{AP75}$\uparrow$ &
      \textit{mAP}$\uparrow$ & \textit{AP50}$\uparrow$ & \textit{AP75}$\uparrow$ \\
    \cline{1-7}
   \rowcolor{light-gray0}   M=1,N=3 &\textbf{0.757} &\textbf{0.913} &\textbf{0.582} &0.397 &0.408 &0.303 \\
      M=3,N=3 &0.728 &0.870 &0.567 &0.465 &0.491 &0.425 \\
    \rowcolor{light-gray0}  M=5,N=1 &0.745 &0.752 &0.501 &0.352 &0.369 &0.315 \\
      \textbf{M=5,N=3} &0.736 &0.879 &0.574 &\textbf{0.482} &\textbf{0.502} &\textbf{0.439} \\
   \rowcolor{light-gray0}   M=5,N=5 &0.734 &0.868 &0.566 &0.474 &0.496 &0.413\\
      M=5,N=10 &0.732 &0.868 &0.560 &0.465 &0.493 &0.422 \\
    \cline{1-7}
    \end{tabular}%
    }
    \caption{Ablation on different settings on $M,N$ pairs. The setting in \textbf{bold} is our primary setting. When $M=3$, we select the \textit{$10$th}, \textit{$20$th}, and the final layers; when \(M=1\), we select the final layer.}
    \label{tab:hyper}
\end{table}

We also perform the \textit{S2A} generalization technique ablation. We retain all hyperparameters and settings but discard non-distortion class predictions. The ablation results are recorded in Tab.~\ref{tab:S2A}. The \textit{S2A} technique enhances the model's generalization ability in real-world scenarios.

Additionally, we perform detailed ablation studies on various hyperparameter settings. First, we explore different combinations of \(M\) and \(N\) with results shown in Tab.~\ref{tab:hyper}. We also conduct an ablation analysis on the selection of \(M=5\) detector positions. For \(M = 5\), we test two  configurations for comparison: early layers (``\textit{Early}"), selecting layers $2$, $4$, $6$, $8$, and $10$; last layers (``\textit{Last}"), selecting the final $5$ layers. The results are presented in Tab.~\ref{tab:position}.  When the number of \(M\) and \(N\) is equal to or slightly greater than the maximum number of target regions in the training images, the detection model achieves better performance. Furthermore, selecting LLM layers in a uniformly distributed order by layer number (our setting) is most beneficial for improving performance.

\begin{table}[!t]\small
    \centering
    \renewcommand\arraystretch{1.1}
    \setlength{\tabcolsep}{5pt}
    \belowrulesep=0pt\aboverulesep=0pt

    \resizebox{\linewidth}{!}{%
    \begin{tabular}{l|ccc|ccc}
    \hline
    \multirow{2}{*}{\textbf{Version}}
      & \multicolumn{3}{c|}{\textbf{Synthetic}}
      & \multicolumn{3}{c}{\textbf{Authentic}}\\

    \cline{2-7}
      &\textit{mAP}$\uparrow$ & \textit{AP50}$\uparrow$ & \textit{AP75}$\uparrow$ &
      \textit{mAP}$\uparrow$ & \textit{AP50}$\uparrow$ & \textit{AP75}$\uparrow$ \\
    \cline{1-7}
   \rowcolor{light-gray0}   \textit{Ours}&0.736 &0.879 &0.574 &\textbf{0.482} &\textbf{0.502} &\textbf{0.439} \\
      \textit{Early}&0.732 &0.874 &0.580 &0.368 &0.374 &0.301 \\
    \rowcolor{light-gray0}  \textit{Last} &\textbf{0.743} &\textbf{0.904} &\textbf{0.590} &0.421 &0.417 &0.349\\
    \cline{1-7}
    \end{tabular}%
    }
    \caption{Ablation on different layer selection strategies.}
    \label{tab:position}
\end{table}

\section{Conclusion}

We propose the \textbf{VIGIL}, an LMM-based framework for spatial visual distortion detection. We construct the \textbf{VIGIL-140K} through rigorous visual quality filtering and spatial distortion combination strategies within a source image pool of over $1000K$ samples. 
During training, we utilize different LLM layers synchronous detection, thereby fully leveraging the hierarchically enhanced image features. In the inference stage, we retain the features of areas predicted as non-distortion classes to alleviate the challenge of \textit{FG-BG} ambiguity in \textit{S2A} generalization. The \textit{VIGIL-8B} demonstrates excellent performance in both synthetic distortion detection and OOD \textit{S2A} tasks.  Our work provides compelling insights into fine-grained local characterization of perceptual visual quality.

\bibliography{aaai2026}

\clearpage
\newpage
\appendix
\begin{table*}[!t]
 \renewcommand\arraystretch{1.12}
\renewcommand\tabcolsep{6pt}
\belowrulesep=0pt\aboverulesep=0pt
\centering
\caption{
Details of the model structure and hyperparameters for the \textit{VIGIL} model.  
}
\vspace{-10pt}
\resizebox{\linewidth}{!}{
\begin{tabular}{l| c| c}
\toprule
\textbf{Model Structure/Training Hyper-Parameters} &  \textbf{Name/Value} &  \textbf{More Information}  \\
\midrule
Vision encoder \textit{init.} & \textit{InternViT-300M-448px} &\textit{Parameter size=}$304.01$M\\
Vision projector \textit{init.} & \textit{2-layers MLP+GeLU}&\textit{Parameter size}=$27.54$M \textit{(Layernorm+Linear(1024,3584)+GELU+Linear(3584,3584))} \\
LLM \textit{init.} & \textit{Qwen-2.5-7B} &\textit{parameter size}=$7612.82$M,Decoder-only model\\

Classifier (for each detector)& \textit{Linear ($3584,9$)} &\textit{parameter size}=$0.03$M\\

Regressor (for each detector)& \textit{Linear ($3584,4$)} &\textit{parameter size}=$0.01$M\\

Image Token Feature Dimension (hidden size)         & $3584$&/\\

Batch Size &$16$ &\textit{Per device train batch size=$2$ (for pair-wise training, this is set to $1$)} \\
LR Max & 2e-5 &/. \\
Gradient Accumulation Steps   & $2$&/ \\
Numerical Precision      & $\mathtt{bfloat16}$&/ \\
Epoch & $1$ & / \\
Validation Steps & $500$ & / \\
Optimizer & AdamW&/ \\
Activation Checkpointing &\checkmark&/ \\
Deepspeed Stage & $2$ &/ \\
\bottomrule
\end{tabular}
}

\label{tab:model_hyperparam}

\end{table*}
\section{Metrics Explanation}
We provide the definitions and formulas for commonly used metrics in distorted area detection: mean Average Precision (\textit{mAP}), Average Precision at specific Intersection-over-Union (\textit{IoU}) thresholds (\textit{AP50}, \textit{AP75}), and Generalized Intersection over Union (\textit{GIoU}).

\subsection{Average Precision at IoU Threshold 50 (AP50)}
AP50 is the Average Precision calculated at an IoU threshold of 50\%. It measures the accuracy of predicted bounding boxes at this fixed threshold.
\[
\text{AP50} = \frac{\sum_{t} \text{Precision}(t) \cdot \text{Recall}(t)}{\text{Total number of samples at IoU} \geq 0.5}
\]
where \( t \) represents the threshold at which the predictions are considered correct.

\subsection{Average Precision at IoU Threshold 75 (AP75)}
AP75 is similar to AP50, but with a stricter IoU threshold of 75\%. It focuses on evaluating the precision of predicted boxes that closely match the ground truth.
\[
\text{AP75} = \frac{\sum_{t} \text{Precision}(t) \cdot \text{Recall}(t)}{\text{Total number of samples at IoU} \geq 0.75}
\]

\subsection{Mean Average Precision (mAP)}
The mean Average Precision (mAP) is the average of the Average Precision (AP) across all object classes (with thresholds: $[0.05,0.95]$ with the interval $0.05$). It provides a general performance evaluation of how well an object detection model predicts the class and location of objects in images.
\[
\text{mAP} = \frac{1}{C} \sum_{c=1}^{C} \text{AP}_c
\]
where \( C \) is the number of classes, and \( \text{AP}_c \) is the average precision for class \( c \).

\subsection{Generalized Intersection over Union (GIoU)}
The Generalized Intersection over Union (GIoU) extends IoU by accounting for cases where the predicted bounding box is far from the ground truth box. It is calculated as:
\[
\text{GIoU} = \text{IoU} - \frac{|C - (A \cup B)|}{|C|}
\]
where \( A \) and \( B \) are the predicted and ground truth boxes, and \( C \) is the smallest enclosing box containing both \( A \) and \( B \).

\subsection{Region Quality Filtering Metric: Brightness}

The brightness is the average value of the V component in the HSV color space:

\[
B = \frac{1}{N} \sum_{i=1}^{N} V_i
\]
where \( B \) is the brightness, \( V_i \) is the brightness value (V component) of each pixel, and \( N \) is the total number of pixels in the image.

\subsection{Region Quality Filtering Metric: Colorfulness}

The colorfulness is calculated using the method by Hasler \& Süsstrunk. The formula is:
\[
C_\text{std}=\sqrt{\text{std}(R-G)^2 + \text{std}(0.5(R+G)-B)^2}
\]

\[
C_\text{mean}=\sqrt{\text{mean}(R-G)^2 + \text{mean}(0.5(R+G)-B)^2}
\]
\[
C = \frac{C_\text{std} + 0.3 \times C_\text{mean}}{C_{\text{MAX}}}
\]
where \( C \) is the colorfulness, \( R \), \( G \), and \( B \) are the red, green, and blue components of the image, \(\text{std}\) and \(\text{mean}\) are the standard deviation and mean respectively, and \( C_{\text{MAX}}=1.3\times \sqrt{2} \) is the maximum colorfulness value used for normalization.

\subsection{Region Quality Filtering Metric: Sharpness}

The sharpness is computed using the Laplacian variance of the image:

\[
s_{\text{raw}} = \text{Var}\left( \nabla^2 I \right)
\]
where \( s_{\text{raw}} \) is the Laplacian variance and \( \nabla^2 I \) is the Laplacian operator applied to the image \( I \). The sharpness is then mapped to the range [0, 1] using a logarithmic function:

\[
s_{\text{01}} = \frac{\log(1 + s_{\text{raw}})}{\log(1 + s_{\text{max}})}
\]

\[
s_{\text{max}} = \left\lfloor \frac{12000}{\left(\frac{W}{640}\right) \times\left(\frac{H}{640}\right)} \right\rfloor
\]
where $H$ and $W$ are the \textit{height} and \textit{width} of \textbf{the original image where the tested rectangle region comes from}, \( s_{\text{01}} \) is the log-compressed sharpness value, \( s_{\text{raw}} \) is the Laplacian variance, and \( s_{\text{max}} \) is a constant representing the maximum sharpness value.

\subsection{Region Filtering Thresholds}

To ensure all distorted regions are perceptually meaningful and avoid generating invalid samples, we apply quality control metrics to filter candidate regions before distortion application. Each region must satisfy the following criteria:

\begin{itemize}
    \item \textbf{Brightness constraint:} $0.25 \leq B \leq 0.75$, where $B$ is the mean value of the HSV V-channel, preventing over-dark or over-bright regions.
    \item \textbf{Sharpness constraint:} $S_{\text{norm}} \geq 0.9$, where $S_{\text{norm}} = \text{clip}(s_{01}, 0, 1)$ measures image sharpness via Laplacian variance, filtering out already blurry or out-of-focus regions.
    \item \textbf{Colorfullness constraint:} For all evaluated region candidates $C > 0.2$, avoiding overly monotone regions. If the randomly picked distortion type of the region is related to ``saturation", we additionally require $C < 0.8$,  
\end{itemize}
\begin{table}[!t]\tiny
    \centering
    \renewcommand\arraystretch{1}
    \renewcommand\tabcolsep{14pt}
    \belowrulesep=0pt\aboverulesep=0pt

    \caption{Area number generalization test. \textit{Avg Preds} denotes the average qualified predictions on all the images.[Per column: highest in \textbf{bold}.]}
    \vspace{-10pt}
    \resizebox{\linewidth}{!}{%
    \begin{tabular}{l|ccc}
    \hline
    \multirow{2}{*}{\textbf{Version}}
    
      & \multicolumn{3}{c}{\textbf{\textit{Authentic-Supp}}}\\
  
    \cline{2-4}
      &
     \textit{AP50}$\uparrow$ & \textit{AP75}$\uparrow$ & \textit{Avg Preds}$\uparrow$ \\
       \cline{1-4}
      \textit{Ours}&\textbf{0.384} &\textbf{0.211} &\textbf{4.22} \\
      \textit{Text}&0.142 &0.061 &1.78 \\
     \cline{1-4}
    \end{tabular}%
    }
    \label{tab:area}
\end{table}

\begin{table}[!t]\tiny
    \centering
    \renewcommand\arraystretch{1}
    \renewcommand\tabcolsep{5pt}
    \belowrulesep=0pt\aboverulesep=0pt

    \caption{Different parameter frozen strategies ablation.}
    \vspace{-10pt}
    \resizebox{\linewidth}{!}{%
    \begin{tabular}{l|ccc|ccc}
    \hline
    \multirow{2}{*}{\textbf{Version}}
      & \multicolumn{3}{c|}{\textbf{Synthetic}}
      & \multicolumn{3}{c}{\textbf{Authentic}}\\
  
    \cline{2-7}
      &\textit{mAP}$\uparrow$ & \textit{AP50}$\uparrow$ & \textit{AP75}$\uparrow$ &
      \textit{mAP}$\uparrow$ & \textit{AP50}$\uparrow$ & \textit{AP75}$\uparrow$ \\
       \cline{1-7}
      \textit{Ours (all trainable)}&\textbf{0.736} &\textbf{0.879} &0.574&\textbf{0.482} &\textbf{0.502} &\textbf{0.439}\\
      \textit{F-LLM}&0.721 &0.874 &0.568 &0.461 &0.485 &0.419\\
      \textit{F-LMM}&0.685 &0.832 &0.475 &0.352 &0.403&0.342\\
      \textit{LoRA}&0.730 &0.871 &\textbf{0.579} &0.457 &0.483 &0.415\\
     \cline{1-7}
    \end{tabular}%
    }
    \label{tab:freeze}
\end{table}
\section{Model Structure Supplementary Information}
The model structure and key hyperparameter settings  of \textit{VIGIL-8B} are depicted in Tab.~\ref{tab:model_hyperparam}. 

\section{Experiments Supplementary Information}
We also conduct several supplementary experiments on specific points of concern.
\subsection{Justification on Some Key Issues}
Under the loss design of Eq.~\ref{loss} in the main paper, expanding the number of area query tokens to $15$ should result in a lower loss-lower-bound compared to our multi-detector setup (because the regression loss and the classification loss for positive samples with higher assigned weights are matched multiple times in our setup). Therefore, it is valid that the training loss for the multi-detector setup decreases more rapidly in Fig.~\ref{fig:curve} in the main paper.

\subsection{Implementation Details of the Object Detection Baselines}
For all object detection models, during training on the \textit{VIGIL-140K}, we set the training epochs for all models to $100$. We use the same validation set and the early-stop strategy. The training hyperparameters follow the settings specified in the respective original model configuration files.

\subsection{Justification on the Use of In-domain Baselines}
Because most of the in-domain related works mentioned in \textit{Sec.~\ref{in-domain}} are either not open-source or not directly applicable to our primary tasks, the number of available comparison models is limited. Specifically, the situation is as follows:

- \textit{Q-Ground} is a segmentation-based model, so our data is not suitable for training/evaluation on it.

- \textit{Grounding-IQA} only has the paper available for reference, as the model weights and training/testing code are not open-source.

- \textit{ViDA-UGC}'s main contribution lies in its training data and paradigm. However, its training and testing methods still rely on text-based outputs (which we have already compared in our ablation studies), and the tasks corresponding to this model are substantially different from ours, making it unsuitable for direct comparison.
\subsection{Predicted Area Number Generalization}

To explore whether the number of distortion regions in a single training image sample affects the model's ability to predict the number of regions during testing, we annotated an additional $100$ test images, each containing $4-5$ distortion regions (exceeding the maximum number of labeled regions in the training set). Consistent with the ablation study in \textit{Sec.~\ref{discussion}}, we compare the performance of the \textit{VIGIL-8B}  and the text-based SFT variant on this extended test set. We also recorded the average number of valid predicted regions. Results are shown in Tab.~\ref{tab:area}. The results show that due to the \textbf{overfitting} influence of SFT on output contexts, the number of the model’s predicted valid regions is far less than our setting, thus hindering the area number generalization. Our model effectively alleviates this issue, leading to a significant improvement in performance.

\subsection{Ablation Study on Model Parameter Freezing Strategies}

We explore the model performance under different parameter freezing settings. Keeping all other settings the same, we compare three variants of the model: 

- Freezing only the LLM part (denoted as \textit{F-LLM}),

- Freezing the entire LMM part, retaining only the learnable area query tokens and linear classifiers and regressors (denoted as \textit{F-LMM}),

- Performing LoRA fine-tuning on the LLM and linear classifiers and regressors (denoted as \textit{LoRA}).

Results are shown in Tab.~\ref{tab:freeze}.
The results show that even when all LMM parameters are frozen, the model’s performance on in-domain tasks does not significantly decrease, and its performance on the OOD \textit{S2A} task remains at an acceptable level. The performance drop for other model versions was even less noticeable. This demonstrates that our model has a feasible application scenario: it can be trained combined with the proprietary visual quality assessment LMMs to add distortion detection capabilities without significantly affecting its original performance on other tasks. For example, the base model can still use text-generation capabilities for image quality scoring and textual description tasks, and the training methods described above allow the model’s functional scope to be expanded without impairing its original capabilities. This enables the development of a more versatile visual quality evaluation model.
\section{Additional Justification of Key Points}

\paragraph{\textbf{S2A Test Set.}}
The \textit{COCO} dataset was selected because it was originally designed for object detection and covers a wide variety of real-world scenes. These scenes also contain diverse and naturally occurring visual degradations, making them suitable for evaluating distortion detection under challenging conditions. Owing to the substantial cost of manual annotation, together with the strict annotation verification and filtering protocol adopted in this work, the resulting test set contains fewer than $1$K samples.

\paragraph{\textbf{Evaluation Metrics.}}
The primary evaluation metrics are standard average-precision metrics, including \textit{mAP}, \textit{AP50}, and \textit{AP75}, as detailed in \textbf{Supp.~Sec.~A}. These metrics jointly characterize precision and recall across different intersection-over-union thresholds and therefore account for both missed detections and false-positive predictions.

\paragraph{\textbf{Hyperparameter Selection.}}
The post-processing thresholds ($T_{\mathrm{nms}}$) and ($T_{\mathrm{prob}}$) have limited influence in synthetic scenarios, where the confidence scores of predicted boxes are typically high and often exceed $0.95$. In real-world scenarios, performance is more sensitive to these thresholds. To determine them without using the test set, a combinatorial search was conducted on a validation set of $100$ images. Both thresholds were varied from $1$ to $0$ with a step size of $0.05$, and all possible combinations were evaluated. The combination achieving the highest \textbf{mAP} on the validation set was then used for evaluation. This procedure provides a systematic threshold-selection strategy that can be readily applied to common real-world scenarios.

\paragraph{\textbf{Evaluation of LMM Baselines.}}
For general LMMs, relaxing greedy decoding produces only limited changes in detection performance. Object detection requires identifying the region with the highest prediction confidence, which is conceptually consistent with greedy decoding in LMMs, where the highest-probability token is selected for each coordinate prediction. Accordingly, non-greedy decoding is not necessarily better aligned with the conventional detection formulation.

To assess the influence of prompt wording, five semantically equivalent prompts with different phrasings were evaluated. For each image, the prediction with the highest \textit{mIoU} among the five outputs was retained. The resulting performance was close to that obtained with the default prompt, indicating that the evaluation results are not strongly dependent on a particular prompt formulation.

\paragraph{\textbf{Text-Generation Ability.}}
The proposed training scheme remains compatible with conventional text-generation-based LMM supervised fine-tuning. During inference, the distortion-detection module uses only the \textbf{prefill} stage and does not modify the \textbf{decode} stage used for text generation. To examine this compatibility, the model was jointly trained with the quality-assessment instruction-tuning dataset \textit{Q-Instruct-200K}. The jointly trained model retained both distortion-detection capability and text-generation ability, achieving an accuracy of $64.5\%$ on \textit{Q-Bench-Test}, which is comparable to mainstream IQA-LMMs.

As the present work focuses on distortion detection, a comprehensive evaluation of text-generation tasks is outside its scope. Nevertheless, these results indicate that, with an appropriate mixed-training strategy, the proposed detection capability can serve as a complementary enhancement to IQA-LMMs without substantially limiting their general-purpose functionality.

\paragraph{\textbf{Rationale for the S2A Strategy and Post-Processing.}}
The S2A strategy, which uses the second-highest classification probability, and the post-processing thresholds ($T_{\mathrm{nms}}$) and ($T_{\mathrm{prob}}$) are designed to exploit latent information in synthetic data without requiring human annotations. Together, they balance cross-domain information extraction and pseudo-label reliability under a resource-efficient setting.

When training relies exclusively on synthetic distortions, domain generalization becomes a central challenge. The S2A strategy adopts a relatively aggressive information-mining mechanism to extract useful signals from cross-domain data. Such a strategy may also introduce additional noise and false-positive predictions. The subsequent post-processing stage mitigates these effects by filtering predictions according to ($T_{\mathrm{nms}}$) and ($T_{\mathrm{prob}}$). This combination supports domain generalization while avoiding the cost of large-scale manual annotation.

\paragraph{\textbf{Base Model Selection.}}
\textit{InternVL-3-8B} was selected primarily because its fine-tuning pipeline is convenient and can be readily transferred within the InternVL model family. Preliminary experiments compared 8B-scale models from the InternVL-2, InternVL-2.5, InternVL-3, and InternVL-3.5 series and showed comparable performance across these variants. These observations suggest that base-model selection is largely dependent on empirical performance under the specific task and training configuration, rather than solely on model release recency.

\section{Distortion Combination Details}

This section provides detailed mathematical formulations for all $31$ distortion sub-types across $8$ major categories used in our training and in-domain test datasets. As described in the main text, following \textbf{KADIS-700K}, we consider $10$ common distortion types which are merged into $8$ major categories based on perceptual similarity: overexposure is merged with contrast strengthen, and underexposure is merged with contrast weaken. Each distortion is applied at two severity levels: \textsc{noticeable"} and \textsc{severe"}.

To simulate the irregular nature of real-world distortions as closely as possible, we randomly select an irregular region within $1/2$ to $1$ of the ratio of each qualified rectangular distortion area. We ensure that the four-connected region of the distortion overlaps with the four edges of the rectangle. Finally, we apply a closing morphological operation at the edges of the selected distortion area to ensure the integrity of the region.

In the formulations below, $I$ represents the input image and $I_{LQ}$ represents the distorted output. Unless otherwise specified, images are represented as floating-point arrays with pixel values normalized to the range $[0,1]$ (i.e., $I \in [0,1]^{H \times W \times C}$). The oversharpen distortion is an exception that operates directly on uint8 images in the range $[0,255]$.

\subsection{Blur (6 sub-types)}

\noindent\textbf{Motion Blur:}
\begin{equation*}
    I_{LQ} = I \ast K_{\text{motion}}(r, \sigma, \theta)
\end{equation*}
where $(r, \sigma) \in \{(15, 7), (20, 12)\}$, $\theta \sim \mathcal{U}(-90°, 90°)$, and $\ast$ denotes convolution.

\noindent\textbf{Gaussian Blur:}
\begin{equation*}
    I_{LQ} = \mathcal{G}_{\sigma}(I), \quad \sigma \in \{3.6, 6.0\}
\end{equation*}
where $\mathcal{G}_{\sigma}$ denotes Gaussian filtering with standard deviation $\sigma$.

\noindent\textbf{Glass Blur:}
\begin{equation*}
    I_{LQ} = \mathcal{G}_{\sigma}(\text{ShufflePixels}(\mathcal{G}_{\sigma}(I), s, n))
\end{equation*}
where $(\sigma, s, n) \in \{(1.2, 2, 2), (1.6, 4, 2)\}$, $s$ is shift range, and $n$ is iteration count.

\noindent\textbf{Lens Blur:}
\begin{equation*}
    I_{LQ}^{(c)} = I^{(c)} \ast K_{\text{disk}}(r), \quad c \in \{R,G,B\}
\end{equation*}
where $r \in \{4, 8\}$ is the disk kernel radius.

\noindent\textbf{Zoom Blur:}
\begin{equation*}
    I_{LQ} = \frac{1}{|\mathcal{Z}|+1}\left(I + \sum_{z \in \mathcal{Z}} \text{Zoom}(I, z)\right)
\end{equation*}
where $\mathcal{Z} \in \{\text{arange}(1, 1.10, 0.02), \text{arange}(1, 1.21, 0.02)\}$.

\noindent\textbf{Jitter Blur:}
\begin{equation*}
    I_{LQ} = \text{ShufflePixels}(I, s, 1), \quad s \in \{3, 5\}
\end{equation*}

\subsection{Noise (6 sub-types)}

\noindent\textbf{Gaussian Noise (RGB):}
\begin{equation*}
    I_{LQ} = \text{clip}(I + \mathcal{N}(0, \sigma^2), 0, 1), \quad \sigma \in \{0.15, 0.25\}
\end{equation*}

\noindent\textbf{Gaussian Noise (YCrCb):}
\begin{equation*}
    I_{LQ} = \text{YCrCb2RGB}(Y + \mathcal{N}_Y, Cr + \mathcal{N}_{Cr}, Cb + \mathcal{N}_{Cb})
\end{equation*}
where $(Y, Cr, Cb) = \text{RGB2YCrCb}(I)$, with $(\sigma_Y, \sigma_{Cr}, \sigma_{Cb}) \in \{(0.07, 0.133, 0.133), (0.09, 0.252, 0.252)\}$.

\noindent\textbf{Speckle Noise:}
\begin{equation*}
    I_{LQ} = \text{clip}(I + I \odot \mathcal{N}(0, s^2), 0, 1), \quad s \in \{0.28, 0.42\}
\end{equation*}

\noindent\textbf{Spatially Correlated Noise:}
\begin{equation*}
    I_{LQ} = \text{Blur}_{3 \times 3}(I + \mathcal{N}(0, \sigma^2)), \quad \sigma \in \{0.14, 0.22\}
\end{equation*}

\noindent\textbf{Poisson Noise:}
\begin{equation*}
    I_{LQ} = \frac{\text{Poisson}(\lambda I)}{\lambda}, \quad \lambda \in \{40, 15\}
\end{equation*}

\noindent\textbf{Impulse Noise (Salt \& Pepper):}
\begin{equation*}
    I_{LQ} = \text{SaltPepper}(I, p), \quad p \in \{0.05, 0.10\}
\end{equation*}
where $p$ is the fraction of affected pixels.

\subsection{Compression (2 sub-types)}

\noindent\textbf{JPEG Compression:}
\begin{equation*}
    I_{LQ} = \text{JPEG}^{-1}(\text{JPEG}(I, q)), \quad q \in \{10, 3\}
\end{equation*}
where $q$ is the quality parameter.

\noindent\textbf{JPEG2000 Compression:}
\begin{equation*}
    I_{LQ} = \text{JP2K}^{-1}(\text{JP2K}(I, q_{\text{dB}})), \quad q_{\text{dB}} \in \{26, 23\}
\end{equation*}
where $q_{\text{dB}}$ is the quality in decibels.

\subsection{Overexposure \& Contrast Strengthen (6 sub-types)}

\noindent\textbf{Brighten Shift (HSV):}
\begin{equation*}
    I_{LQ} = \text{HSV2RGB}(H, S, V + \Delta V)
\end{equation*}
where $(H,S,V) = \text{RGB2HSV}(I)$, $\Delta V \in \{0.39, 0.65\}$.

\noindent\textbf{Brighten Shift (RGB):}
\begin{equation*}
    I_{LQ} = \text{clip}(I + \Delta I, 0, 1), \quad \Delta I \in \{0.325, 0.52\}
\end{equation*}

\noindent\textbf{Brighten Gamma (HSV):}
\begin{equation*}
    I_{LQ} = \text{HSV2RGB}(H, S, V^{\gamma})
\end{equation*}
where $\gamma \in \{0.3375, 0.165\}$.

\noindent\textbf{Brighten Gamma (RGB):}
\begin{equation*}
    I_{LQ} = I^{\gamma}, \quad \gamma \in \{0.45, 0.30\}
\end{equation*}

\noindent\textbf{Contrast Strengthen (Scale):}
\begin{equation*}
    I_{LQ} = \text{ContrastEnhance}(I, f), \quad f \in \{5.2, 8.0\}
\end{equation*}
where $f > 1$ increases contrast.

\noindent\textbf{Contrast Strengthen (Stretch):}
\begin{equation*}
    I_{LQ} = \frac{1}{1 + \left(\frac{\bar{I}}{I + \epsilon}\right)^{f}}, \quad f \in \{12.0, 20.0\}
\end{equation*}
where $\bar{I} = \text{mean}(I)$ and $\epsilon = 10^{-12}$.

\subsection{Underexposure \& Contrast Weaken (6 sub-types)}

\noindent\textbf{Darken Shift (HSV):}
\begin{equation*}
    I_{LQ} = \text{HSV2RGB}(H, S, V - \Delta V), \quad \Delta V \in \{0.30, 0.45\}
\end{equation*}

\noindent\textbf{Darken Shift (RGB):}
\begin{equation*}
    I_{LQ} = \text{clip}(I - \Delta I, 0, 1), \quad \Delta I \in \{0.25, 0.40\}
\end{equation*}

\noindent\textbf{Darken Gamma (HSV):}
\begin{equation*}
    I_{LQ} = \text{HSV2RGB}(H, S, V^{\gamma}), \quad \gamma \in \{3.51, 5.2\}
\end{equation*}
where $(H, S, V) = \text{RGB2HSV}(I)$.

\noindent\textbf{Darken Gamma (RGB):}
\begin{equation*}
    I_{LQ} = I^{\gamma}, \quad \gamma \in \{3.38, 4.68\}
\end{equation*}

\noindent\textbf{Contrast Weaken (Scale):}
\begin{equation*}
    I_{LQ} = \text{ContrastEnhance}(I, f), \quad f \in \{0.27, 0.09\}
\end{equation*}
where $f < 1$ reduces contrast.

\noindent\textbf{Contrast Weaken (Stretch):}
\begin{equation*}
    I_{LQ} = \frac{1}{1 + \left(\frac{\bar{I}}{I + \epsilon}\right)^{f}}, \quad f \in \{0.42, 0.30\}
\end{equation*}
where $\bar{I} = \text{mean}(I)$ and $\epsilon = 10^{-12}$.

\subsection{Saturate Strengthen (2 sub-types)}

\noindent\textbf{Strengthen Saturation (HSV):}
\begin{equation*}
    I_{LQ} = \text{HSV2RGB}(H, \text{clip}(f \cdot S, 0, 255), V), \quad f \in \{18.0, 96.0\}
\end{equation*}
where $(H, S, V) = \text{RGB2HSV}(I)$.

\noindent\textbf{Strengthen Saturation (YCrCb):}
\begin{equation*}
    I_{LQ} = \text{YCrCb2RGB}(Y, 128 + f(Cr - 128), 128 + f(Cb - 128))
\end{equation*}
where $(Y, Cr, Cb) = \text{RGB2YCrCb}(I)$, $f \in \{12.0, 24.0\}$.

\subsection{Saturate Weaken (2 sub-types)}

\noindent\textbf{Weaken Saturation (HSV):}
\begin{equation*}
    I_{LQ} = \text{HSV2RGB}(H, f \cdot S, V), \quad f \in \{0.28, 0.0\}
\end{equation*}
where $(H, S, V) = \text{RGB2HSV}(I)$.

\noindent\textbf{Weaken Saturation (YCrCb):}
\begin{equation*}
    I_{LQ} = \text{YCrCb2RGB}(Y, 128 + f(Cr - 128), 128 + f(Cb - 128))
\end{equation*}
where $(Y, Cr, Cb) = \text{RGB2YCrCb}(I)$, $f \in \{0.14, 0.0\}$.

\subsection{Oversharpen (1 sub-type)}

\noindent\textbf{Oversharpen:}
\begin{equation*}
    I_{LQ} = \text{clip}((1 + \alpha) I - \alpha \mathcal{G}_{\sigma}(I), 0, 255)
\end{equation*}
where $\alpha \in \{6.0, 12.0\}$, $\sigma = 5$, and input image is in $[0, 255]$ range.





\section{Limitations}
Due to the fact that most in-the-wild UGC image content cannot serve as an effective source for adding distortions, our dataset is still limited in scale, preventing us from verifying the data scaling law. Additionally, due to resource constraints, we have not yet included large-scale annotated real-world scene data in our training set. As a result, there is still room for improvement in the model's performance. These limitations represent key insights and directions for our future research.
\section{Detailed Statistical Information}

We present comprehensive statistical analyses of our datasets, including the training set (\textit{VIGIL-140K}), the synthetic distortion test set, and the authentic distortion test set. We visualize five key statistical distributions for each dataset: image resolution distribution, bounding box resolution distribution, box-to-image area ratio distribution, number of boxes per image distribution, and distortion type distribution. 

\subsection{Training Set Statistics}

The training set \textit{VIGIL-140K} comprises over $145K$ distorted images with more than $205K$ distortion area annotations.

\begin{figure}[H]
    \centering
    \includegraphics[width=0.95\linewidth]{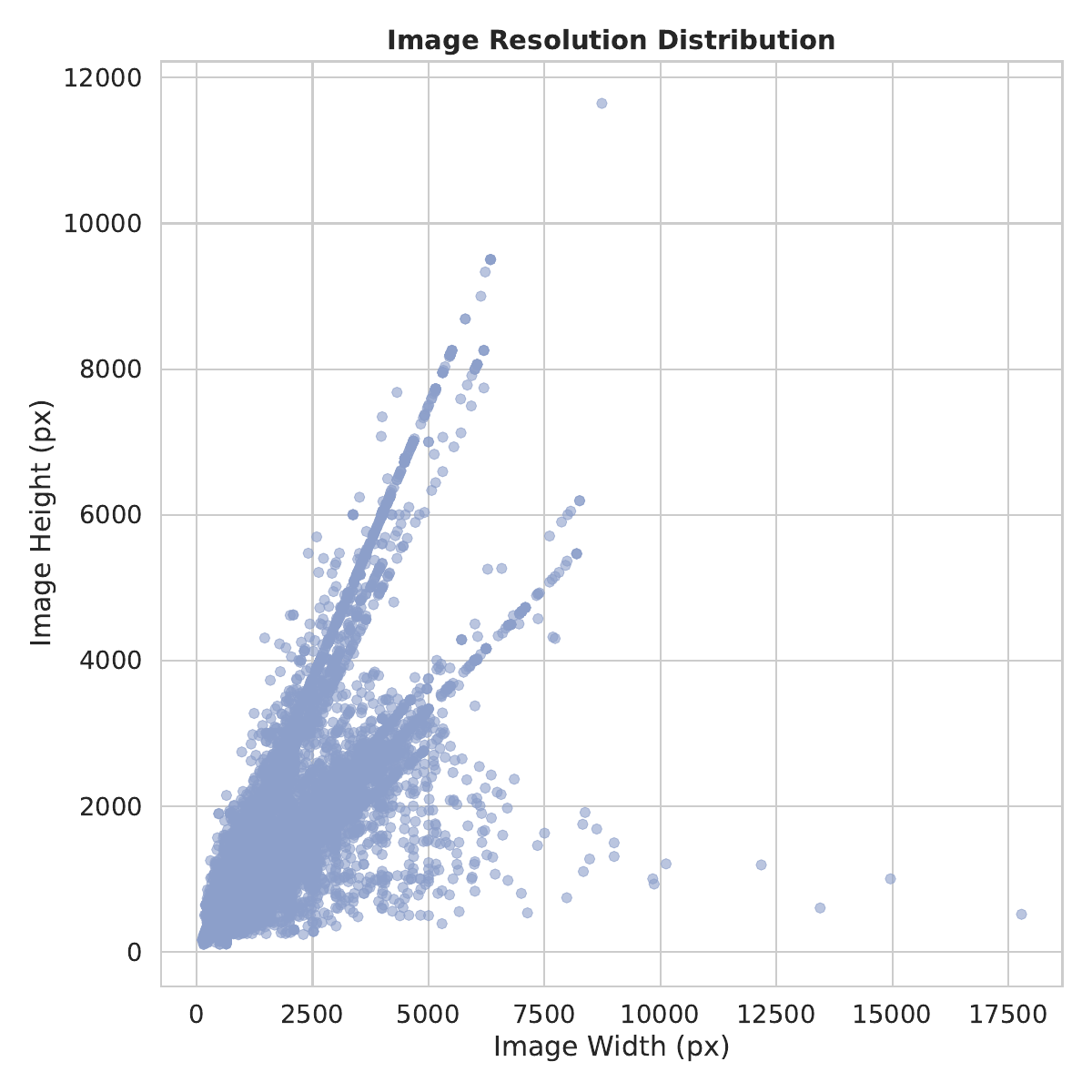}
    \vspace{-5pt}
    \caption{Image resolution distribution of the training set.}
    \label{fig:train_image_res}
\end{figure}
Fig.~\ref{fig:train_image_res} shows the joint distribution of image width and height. The dataset exhibits high diversity, with individual dimensions reaching up to $17,500$ pixels in width and $12,000$ pixels in height. Most images are concentrated within the $500 \times 500$ to $5,000 \times 5,000$ pixel range. Notably, the clear linear patterns suggest that the images follow several standard aspect ratios.

\begin{figure}[H]
    \centering
    \includegraphics[width=0.95\linewidth]{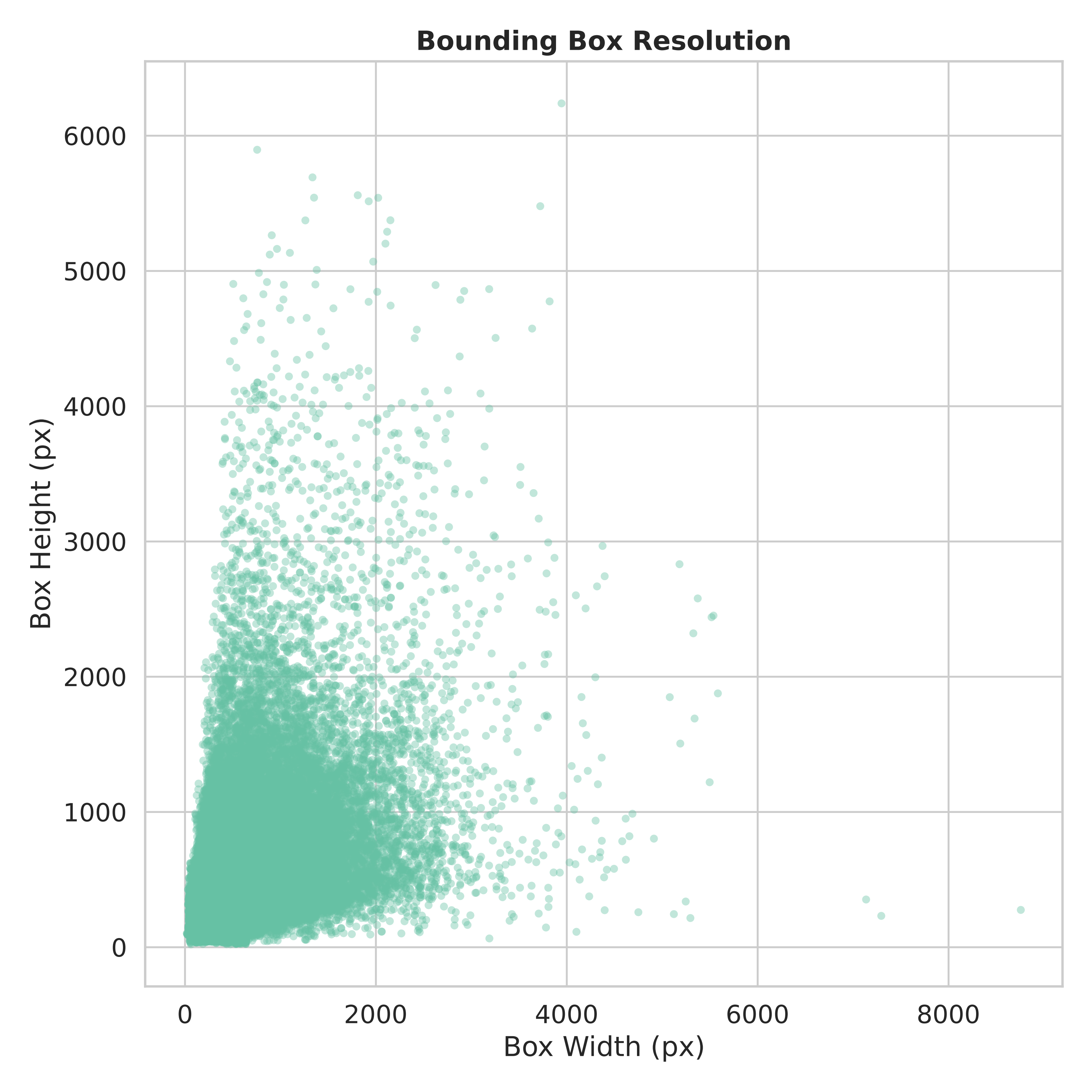}
    \vspace{-5pt}
    \caption{Bounding box resolution distribution of the training set.}
    \label{fig:train_bbox_res}
\end{figure}

Fig.~\ref{fig:train_bbox_res} illustrates the distribution of bounding box resolutions. The distortion regions cover a wide range of spatial scales, contributing to robust model performance across varying region sizes. Most bounding boxes are small to medium (typically below $2{,}000$ pixels), while a substantial number exceed $4{,}000 \times 4{,}000$ pixels.

\begin{figure}[H]
    \centering
    \includegraphics[width=0.95\linewidth]{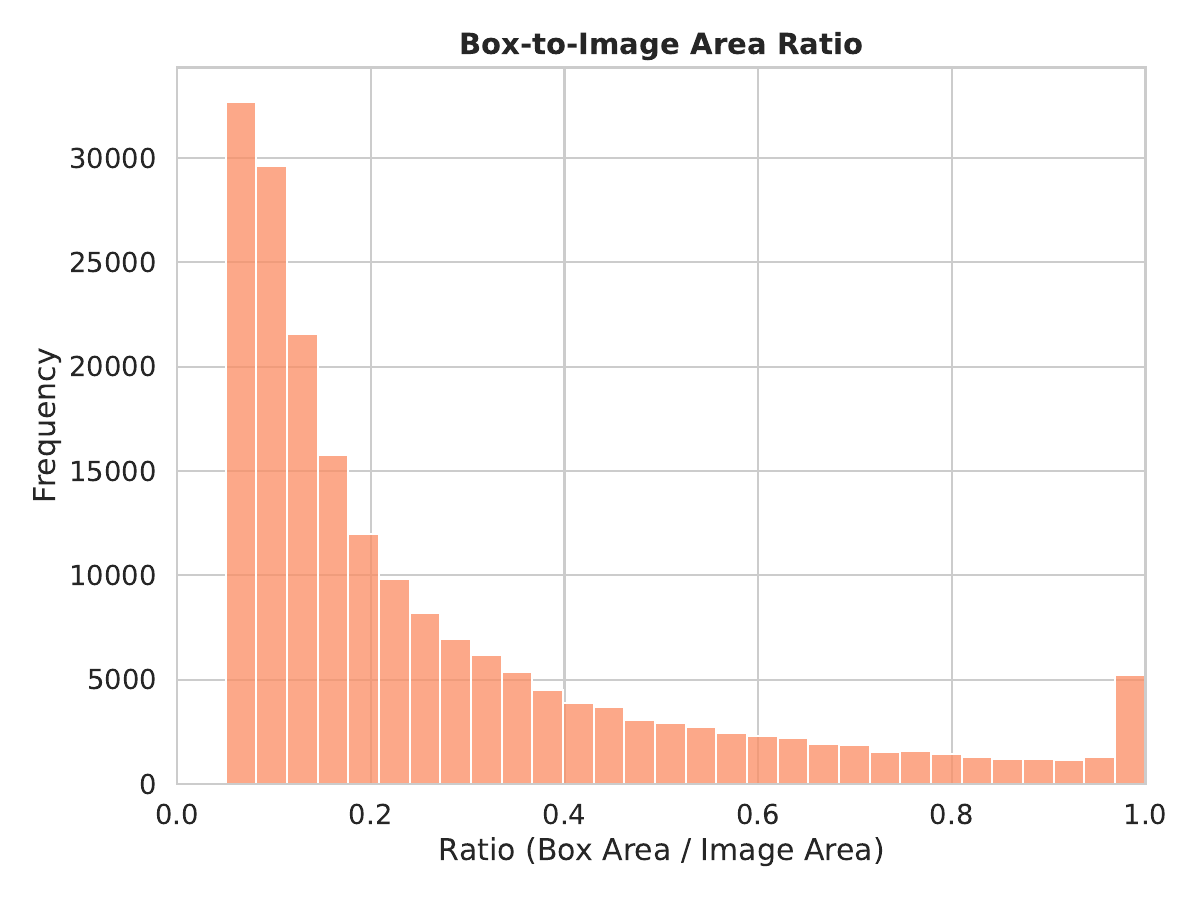}
    \vspace{-5pt}
    \caption{Box-to-image area ratio distribution of the training set.}
    \label{fig:train_ratio}
\end{figure}
The box-to-image area ratio distribution (Fig.~\ref{fig:train_ratio}) demonstrates that most distortion regions occupy between $5\%$ and $30\%$ of the total image area. 

\begin{figure}[H]
    \centering
    \includegraphics[width=0.95\linewidth]{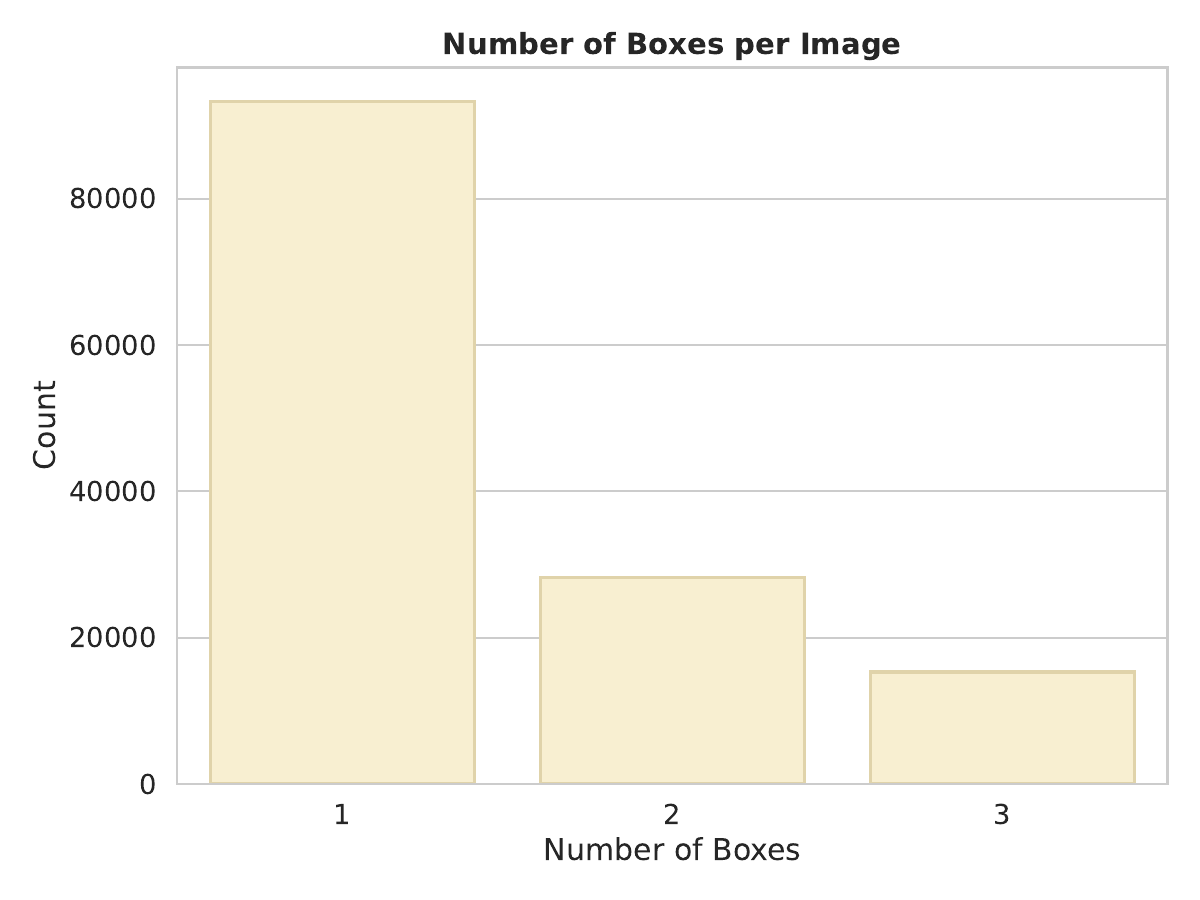}
    \vspace{-5pt}
    \caption{Number of boxes per image distribution of the training set.}
    \label{fig:train_count}
\end{figure}
Fig.~\ref{fig:train_count} shows the distribution of the number of distortion boxes per image. Following our distortion combination algorithm (Algorithm~\ref{distortion_combination}), each image contains $1$--$3$ distortion regions.

\begin{figure}[H]
    \centering
    \includegraphics[width=0.95\linewidth]{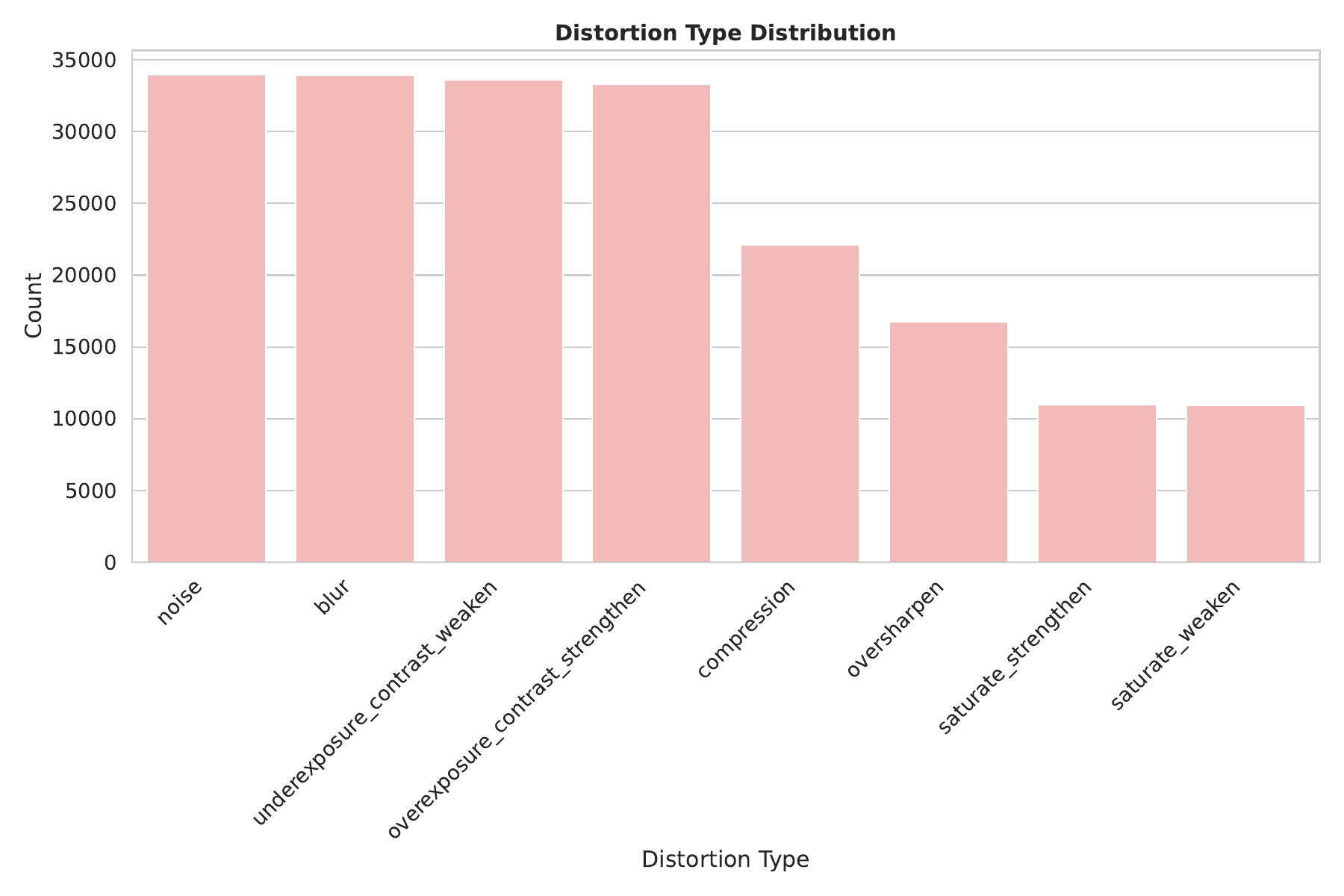}
    \vspace{-5pt}
    \caption{Distortion type distribution (number of labeled areas
 ) of the training set.}
    \label{fig:train_dist_type}
\end{figure}

Fig.~\ref{fig:train_dist_type} presents the distribution of distortion types in the training set, counted by the number of labeled distortion areas. Noise, blur, underexposure/contrast weaken, and overexposure/contrast strengthen are the most prevalent types, each with over 30K instances.
 
\subsection{Synthetic Distortion Test Set Statistics}
The synthetic distortion test set contains 4,600 images generated using the same distortion synthesis pipeline, applied to held-out source images. This test set is used to evaluate the model’s in-domain detection performance.

\begin{figure}[H]
    \centering
    \includegraphics[width=0.95\linewidth]{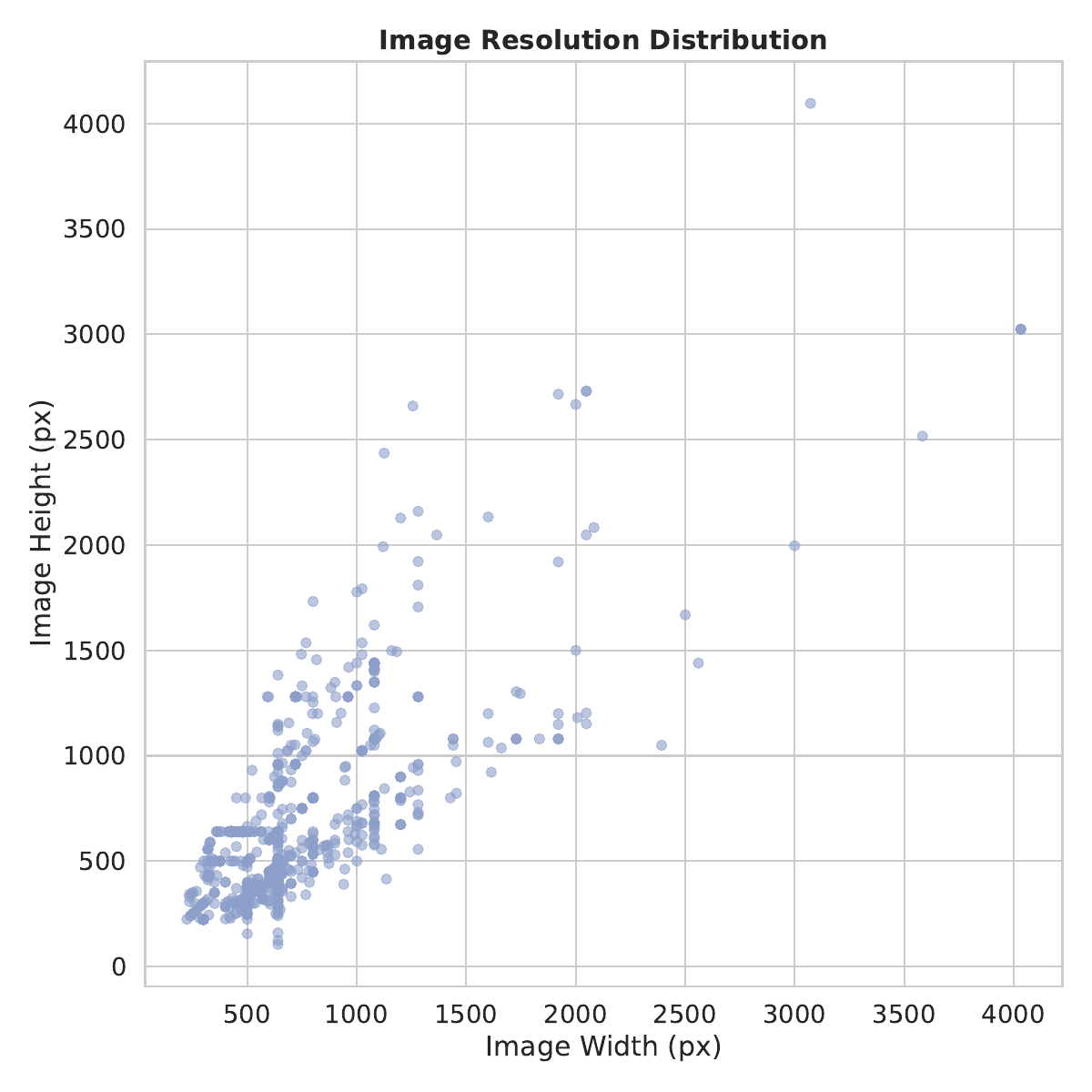}
    \vspace{-5pt}
    \caption{Image resolution distribution of the synthetic distortion test set.}
    \label{fig:syn_image_res}
\end{figure}

\begin{figure}[H]
    \centering
    \includegraphics[width=0.95\linewidth]{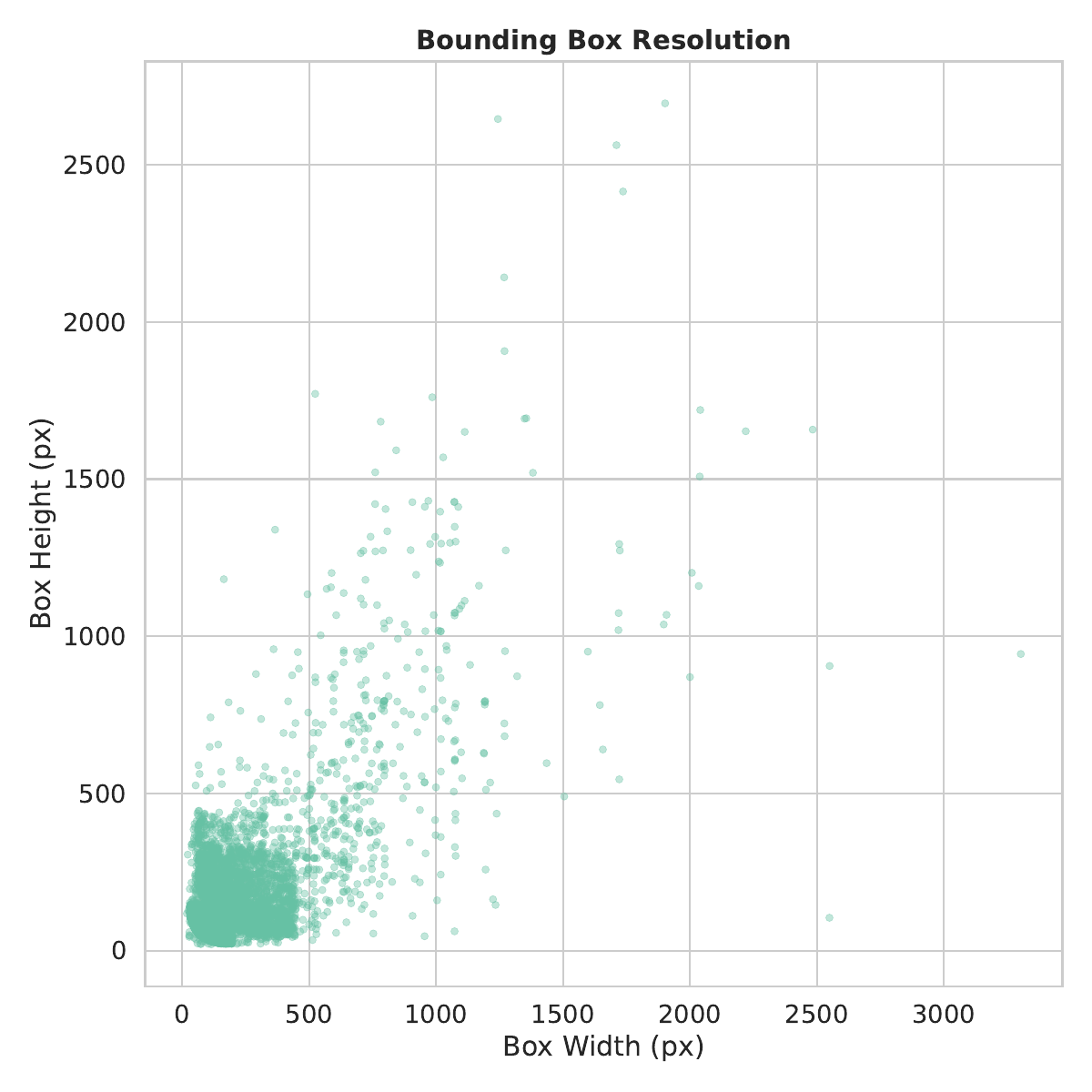}
    \vspace{-5pt}
    \caption{Bounding box resolution distribution of the synthetic distortion test set.}
    \label{fig:syn_bbox_res}
\end{figure}

Figs.~\ref{fig:syn_image_res} and \ref{fig:syn_bbox_res} show the image and bounding box resolution distributions, respectively. The synthetic test set exhibits similar patterns to the training set but with a slightly different distribution due to the different source image pools (\textit{KADIS-700K}, \textit{COCO-2017-test}, and \textit{Unsplash}).

\begin{figure}[H]
    \centering
    \includegraphics[width=0.95\linewidth]{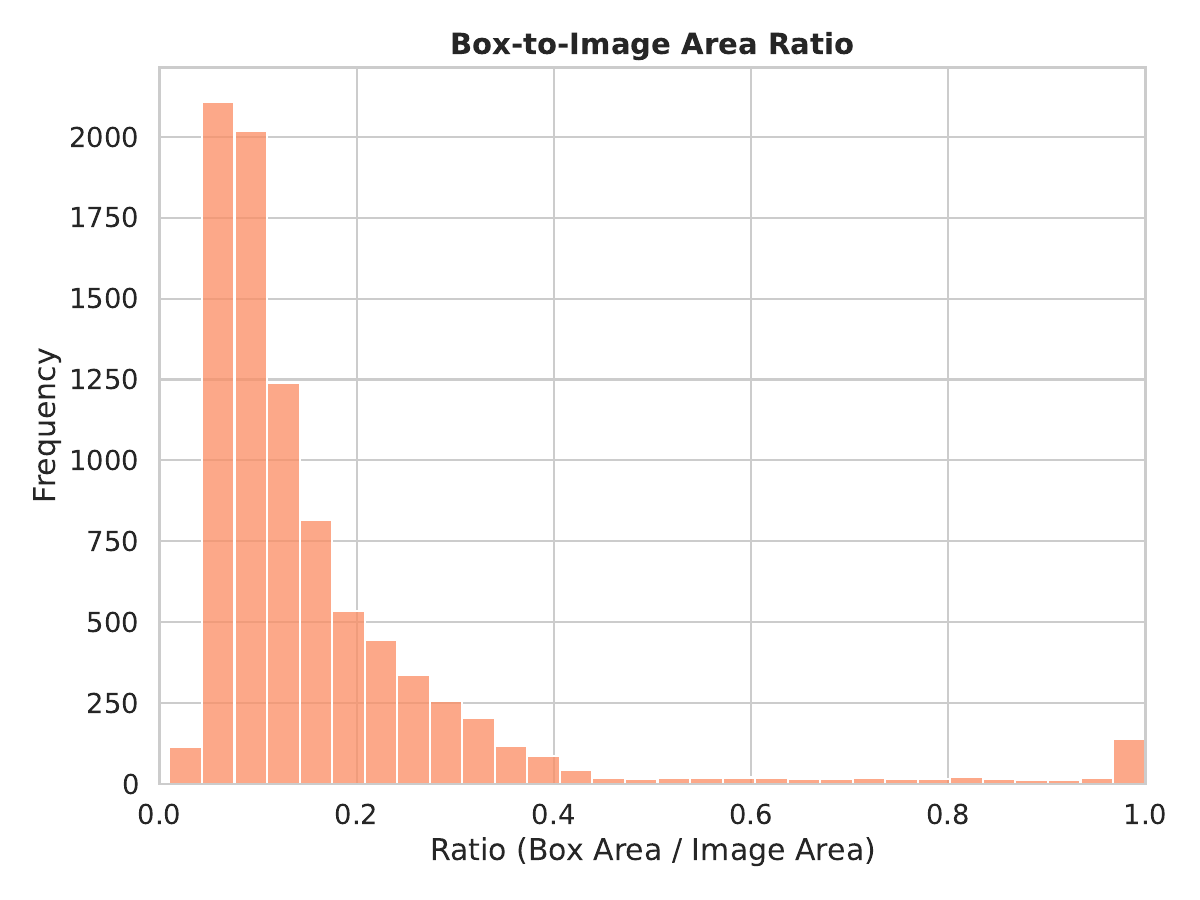}
    \vspace{-5pt}
    \caption{Box-to-image area ratio distribution of the synthetic distortion test set.}
    \label{fig:syn_ratio}
\end{figure}

\begin{figure}[H]
    \centering
    \includegraphics[width=0.95\linewidth]{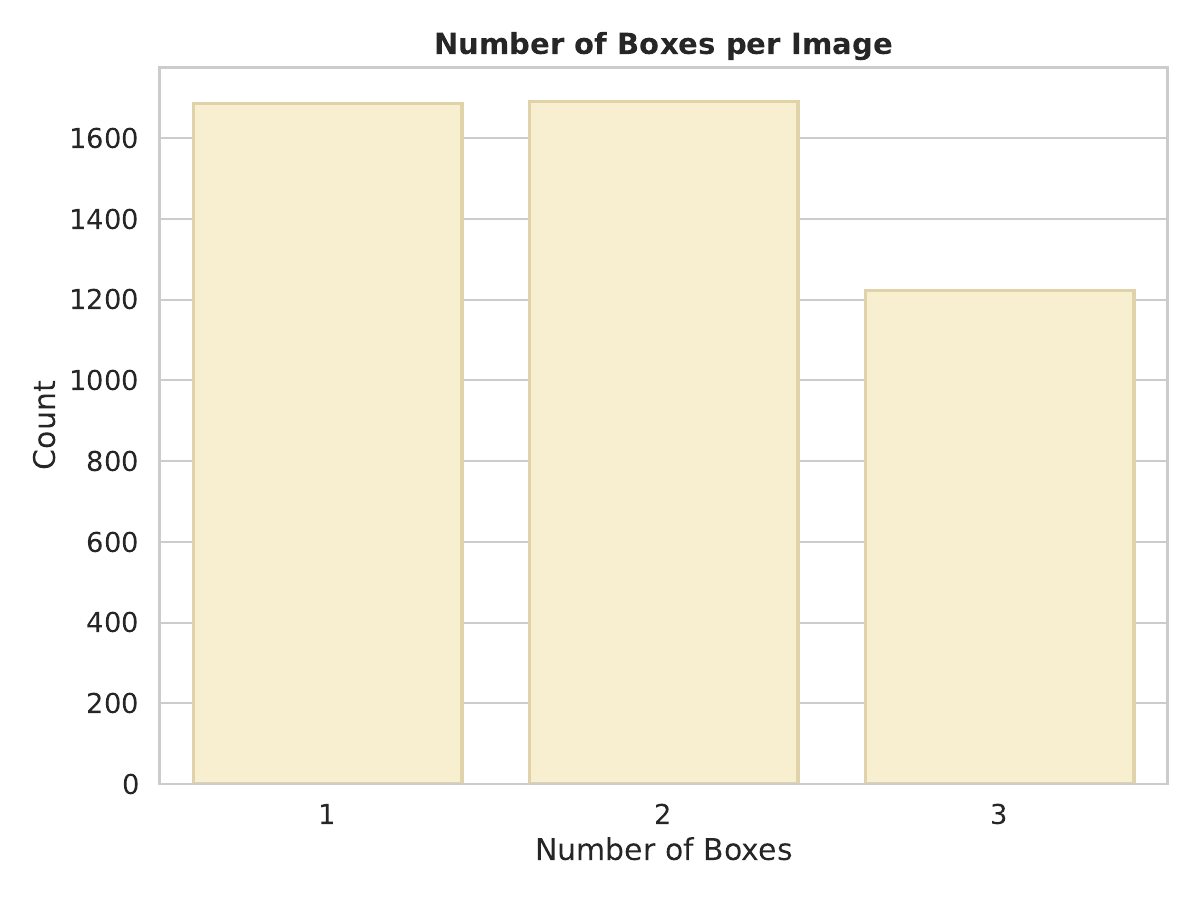}
    \vspace{-5pt}
    \caption{Number of boxes per image distribution of the synthetic distortion test set.}
    \label{fig:syn_count}
\end{figure}

The box-to-image ratio (Fig.~\ref{fig:syn_ratio}) and box count (Fig.~\ref{fig:syn_count}) distributions show the size of the distortion areas relative to the image and the number of distorted regions found in each image within the synthetic test set.

\begin{figure}[H]
    \centering
    \includegraphics[width=0.95\linewidth]{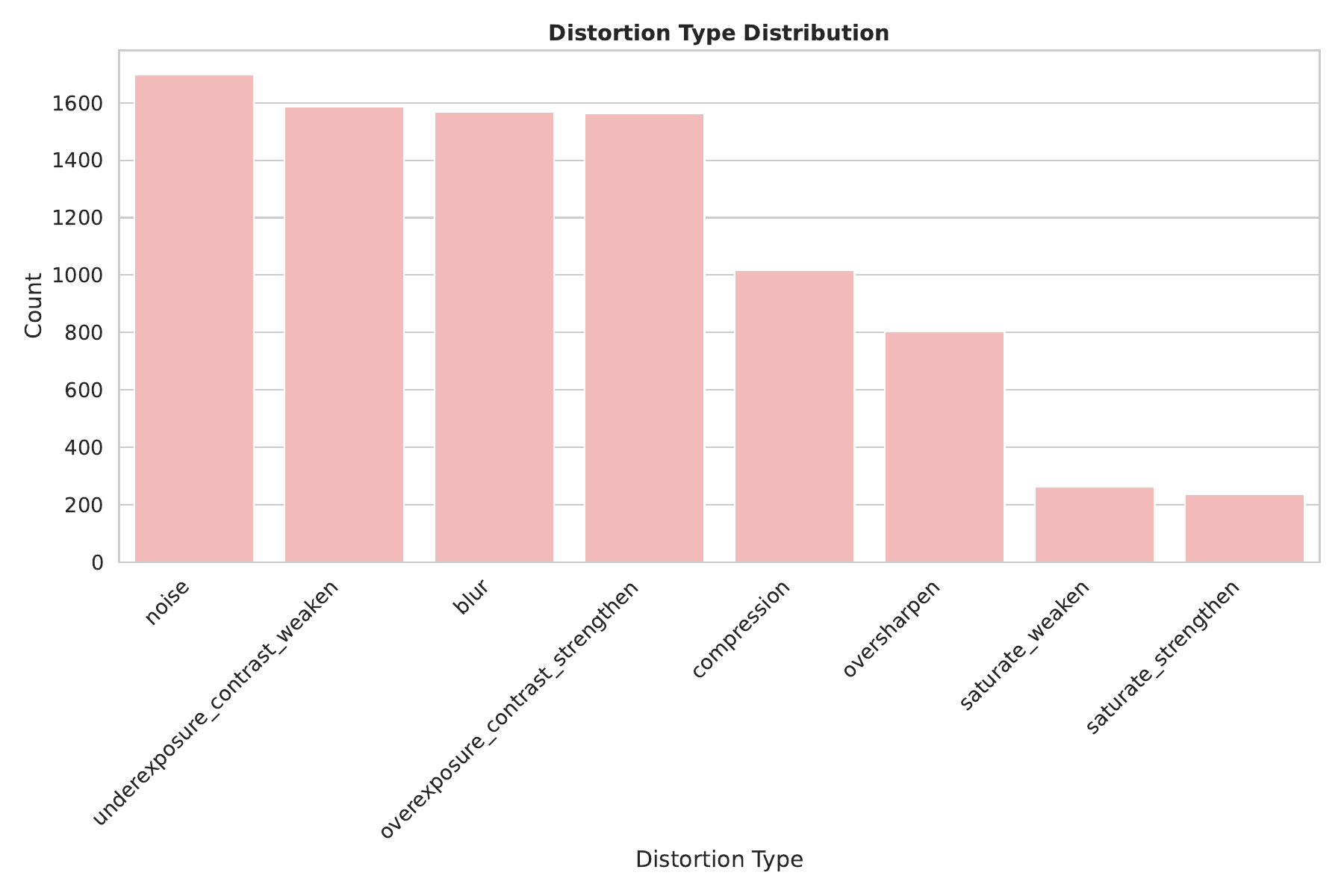}
    \vspace{-5pt}
    \caption{Distortion type distribution of the synthetic distortion test set.}
    \label{fig:syn_dist_type}
\end{figure}

The distortion type distribution (Fig.~\ref{fig:syn_dist_type}) shows a balanced representation across all eight major distortion categories.

\subsection{Authentic Distortion Test Set Statistics}

The authentic distortion test set consists of $815$ human-annotated UGC images with real-world distortions and is used to evaluate out-of-domain (OOD) authentic distortion localization for \textit{S2A} (synthetic-to-authentic) generalization.

\begin{figure}[H]
    \centering
    \includegraphics[width=0.95\linewidth]{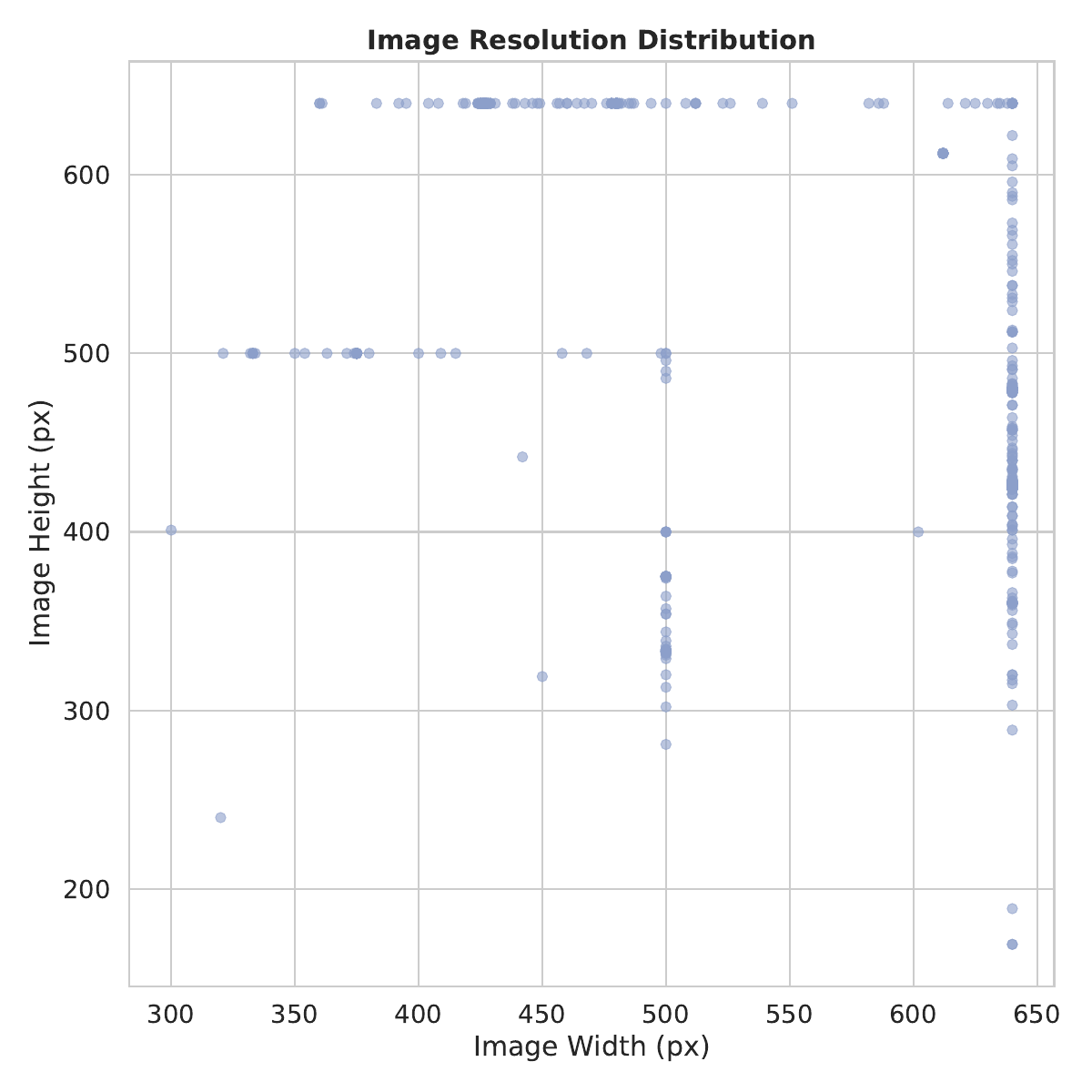}
    \vspace{-5pt}
    \caption{Image resolution distribution of the authentic distortion test set.}
    \label{fig:auth_image_res}
\end{figure}

\begin{figure}[H]
    \centering
    \includegraphics[width=0.95\linewidth]{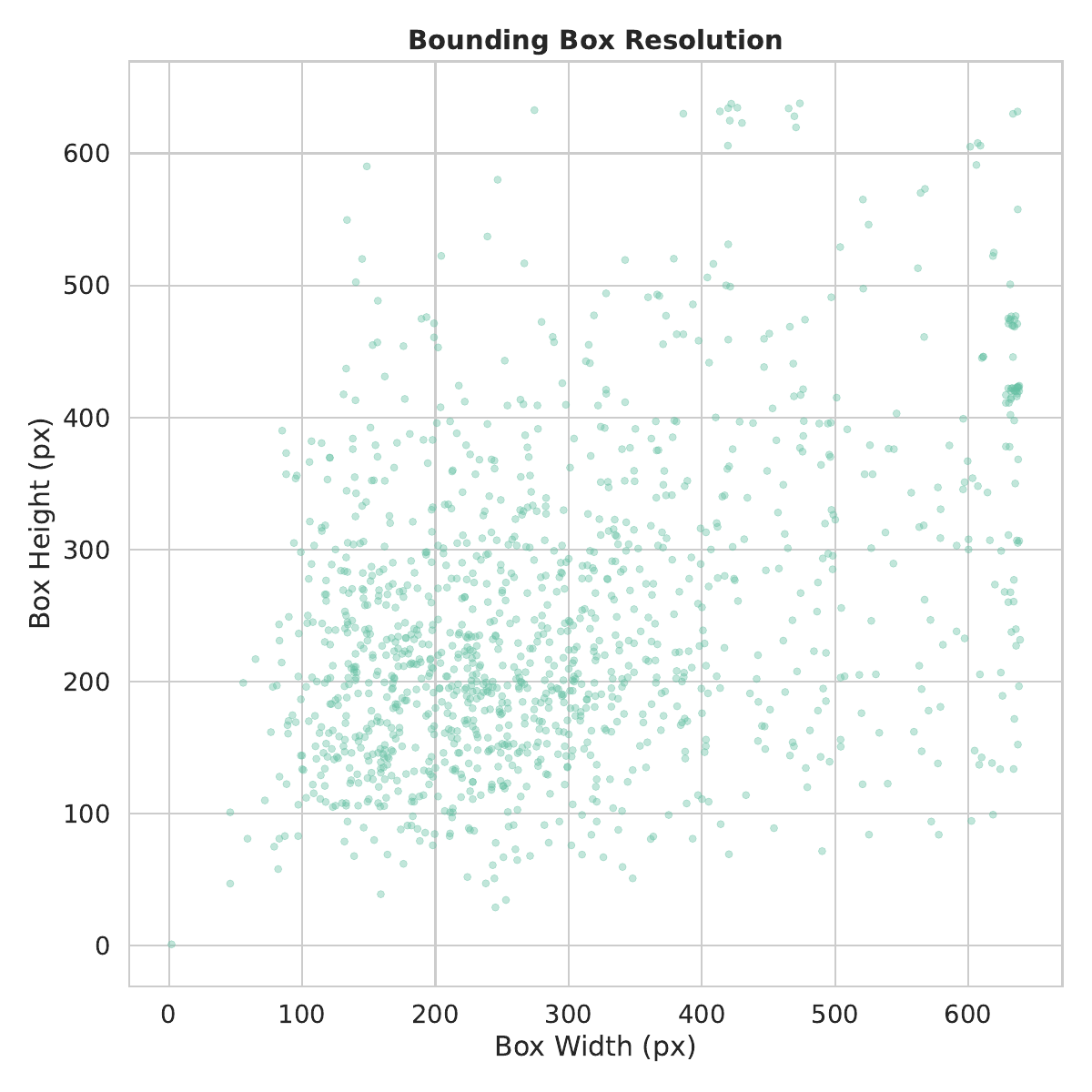}
    \vspace{-5pt}
    \caption{Bounding box resolution distribution of the authentic distortion test set.}
    \label{fig:auth_bbox_res}
\end{figure}

Figs.~\ref{fig:auth_image_res} and \ref{fig:auth_bbox_res} illustrate the resolution distributions for the authentic test set. The image resolutions are mainly concentrated around $500 \times 500$ and $640 \times 640$ pixels, reflecting common resolution scales of UGC images from the \textit{COCO2017} dataset. The bounding box resolutions show a more uniform spread compared to synthetic data, reflecting the irregular and varied nature of authentic distortion regions.

\begin{figure}[H]
    \centering
    \includegraphics[width=0.95\linewidth]{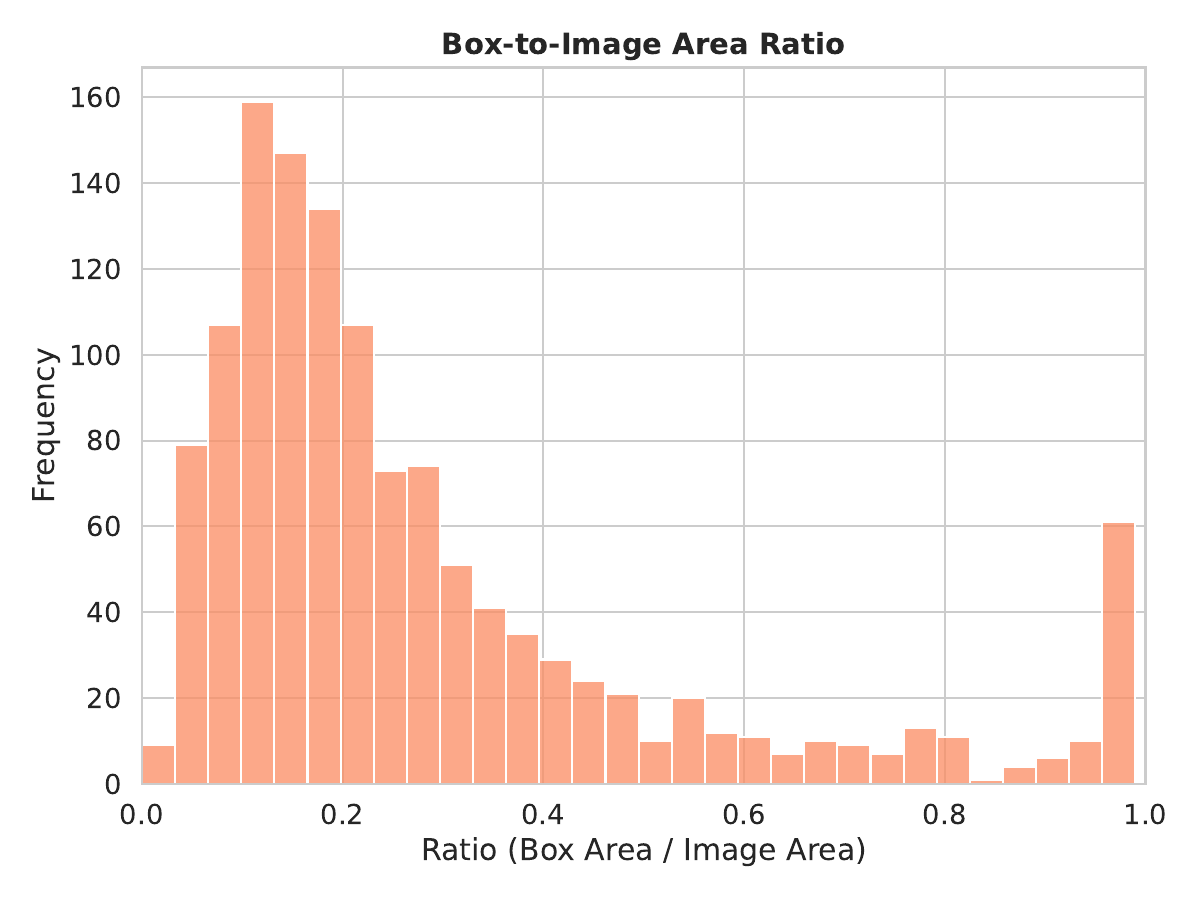}
    \vspace{-5pt}
    \caption{Box-to-image area ratio distribution of the authentic distortion test set.}
    \label{fig:auth_ratio}
\end{figure}

\begin{figure}[H]
    \centering
    \includegraphics[width=0.95\linewidth]{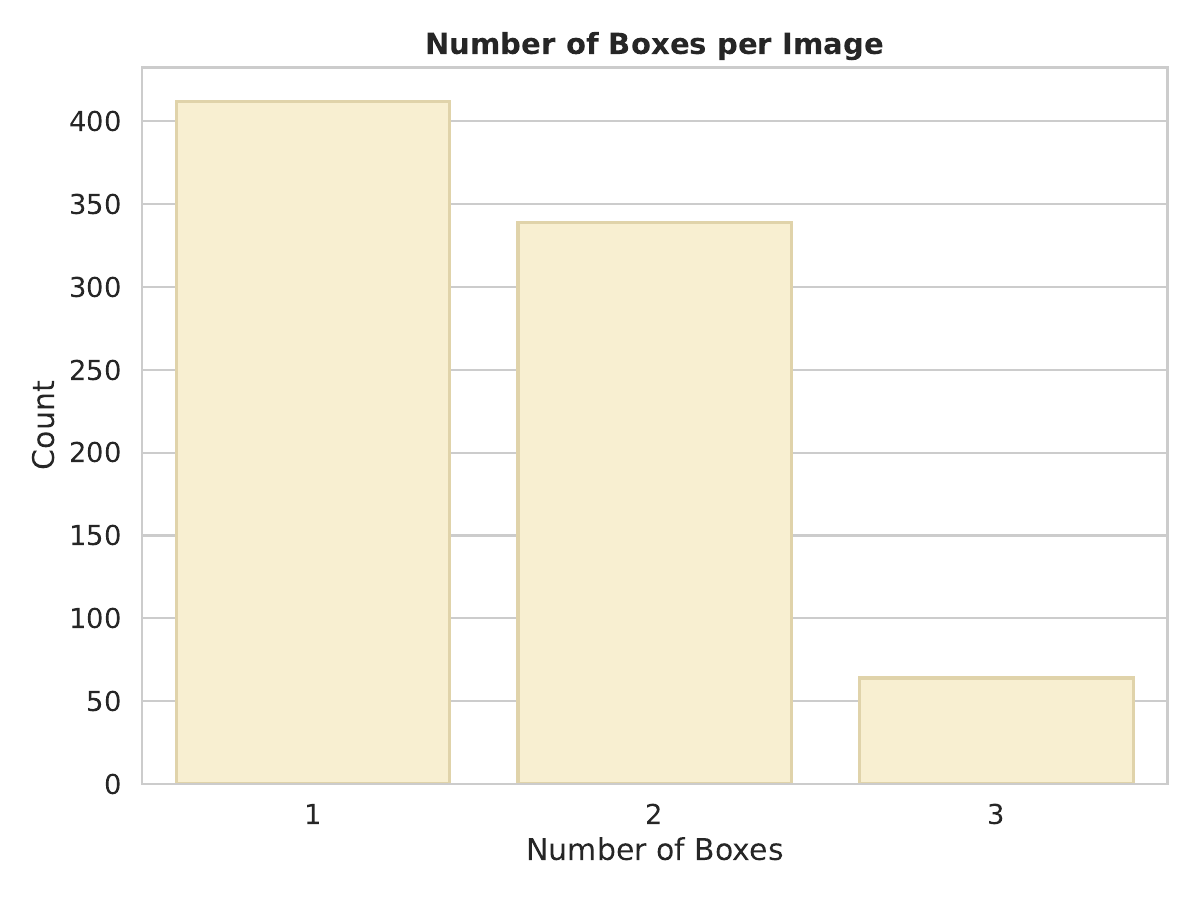}
    \vspace{-5pt}
    \caption{Number of boxes per image distribution of the authentic distortion test set.}
    \label{fig:auth_count}
\end{figure}

The box-to-image area ratio distribution (Fig.~\ref{fig:auth_ratio}) indicates that 
authentic distortions often occupy relatively large portions of the image, with a 
substantial number of regions exceeding 30\% of the image area, while the distribution of distortion region counts per image is summarized in Fig.~\ref{fig:auth_count}.

\begin{figure}[H]
    \centering
    \includegraphics[width=0.95\linewidth]{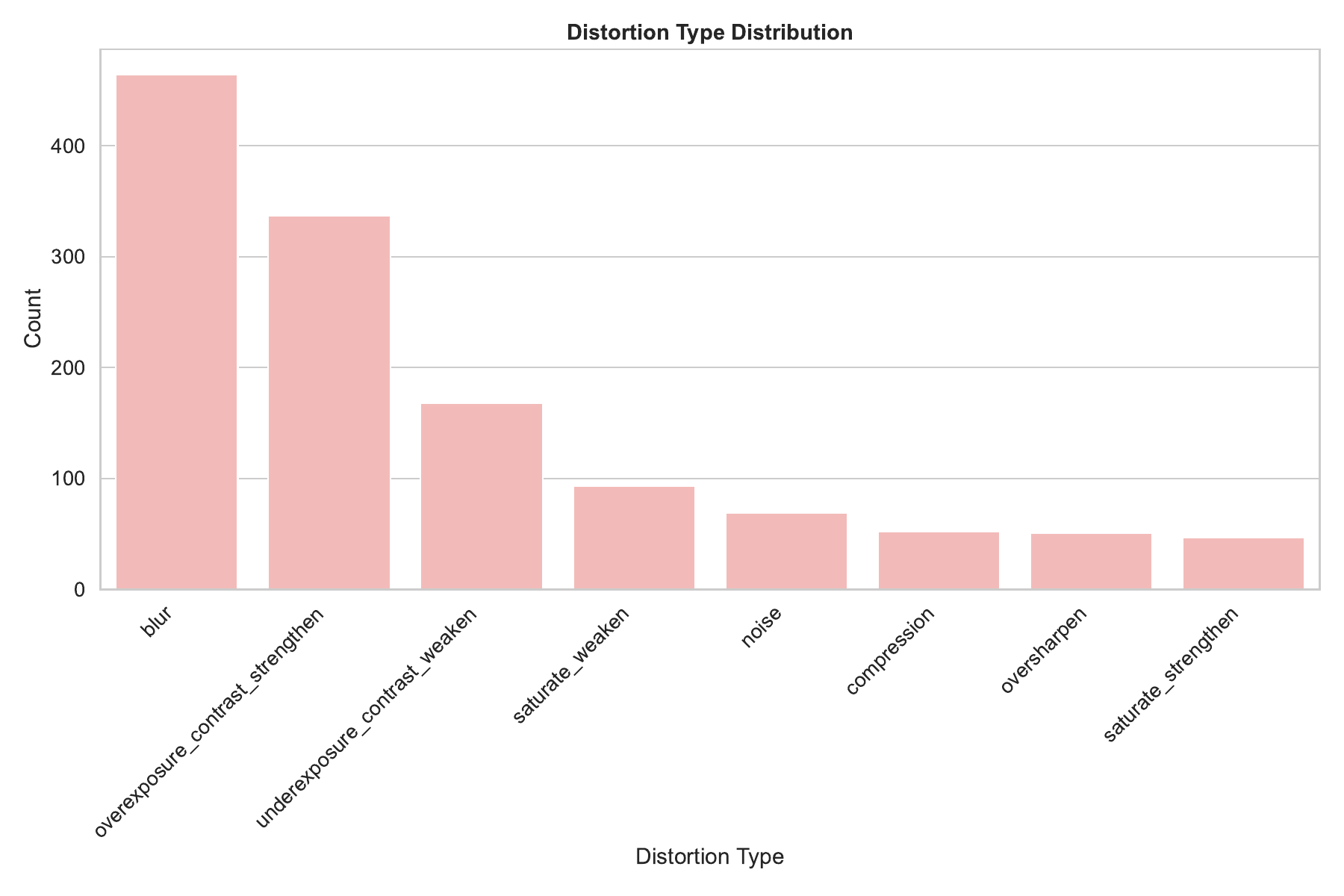}
    \vspace{-5pt}
    \caption{Distortion type distribution of the authentic distortion test set.}
    \label{fig:auth_dist_type}
\end{figure}

Fig.~\ref{fig:auth_dist_type} shows the distribution of authentic distortion types.

\section{Visualization}

\subsection{Training Set Samples}

\begin{figure*}[htbp]
    \centering
    \includegraphics[width=\linewidth]{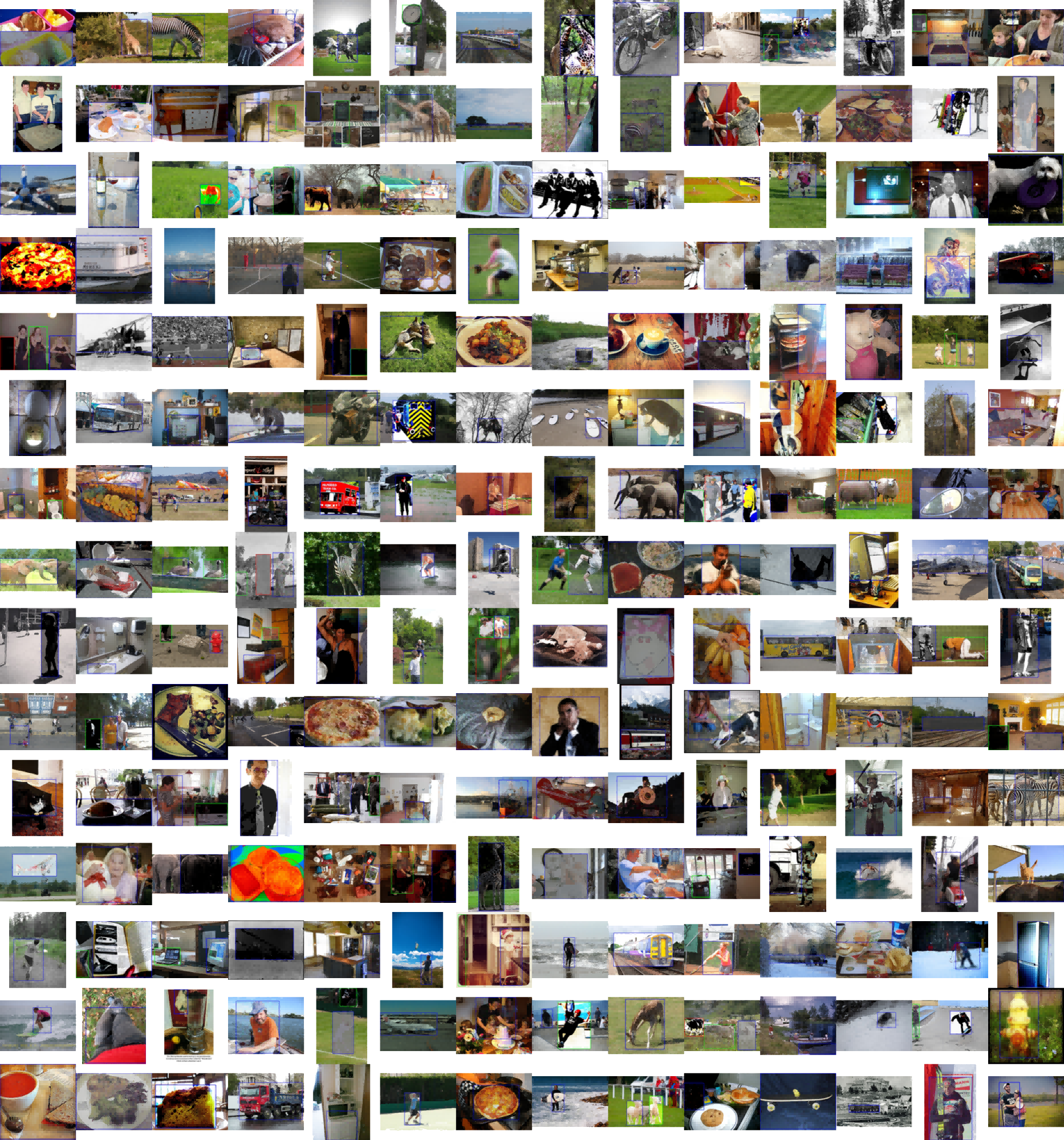}
    \vspace{-7pt}
    \caption{210 representative training samples from \textit{VIGIL-140K}.}
    \label{fig:train_samples}
\end{figure*}

Fig.~\ref{fig:train_samples} shows 210 representative samples from our training set \textit{VIGIL-140K}.

\subsection{Authentic Distortion Test Set Samples}

\begin{figure*}[htbp]
    \centering
    \includegraphics[width=\linewidth]{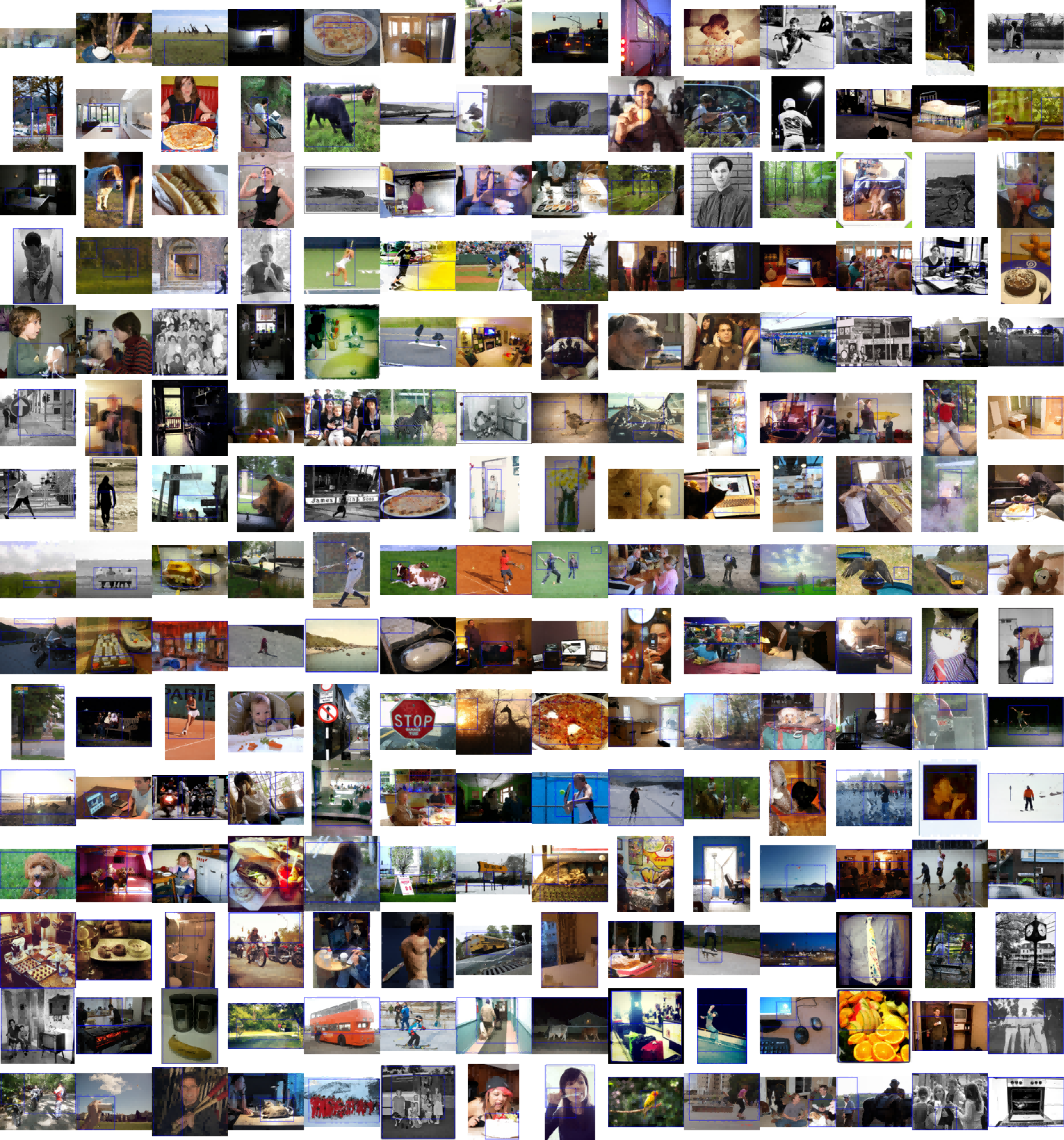}
    \vspace{-7pt}
    \caption{210 samples from the authentic distortion test set.}
    \label{fig:auth_samples}
\end{figure*}

Figs~\ref{fig:auth_samples} presents 210 samples from our authentic distortion test set, showcasing real-world distortions encountered in user-generated content for evaluating OOD generalization.

\subsection{Model Predictions}
\begin{figure*}
    \centering
    \includegraphics[width=\linewidth]{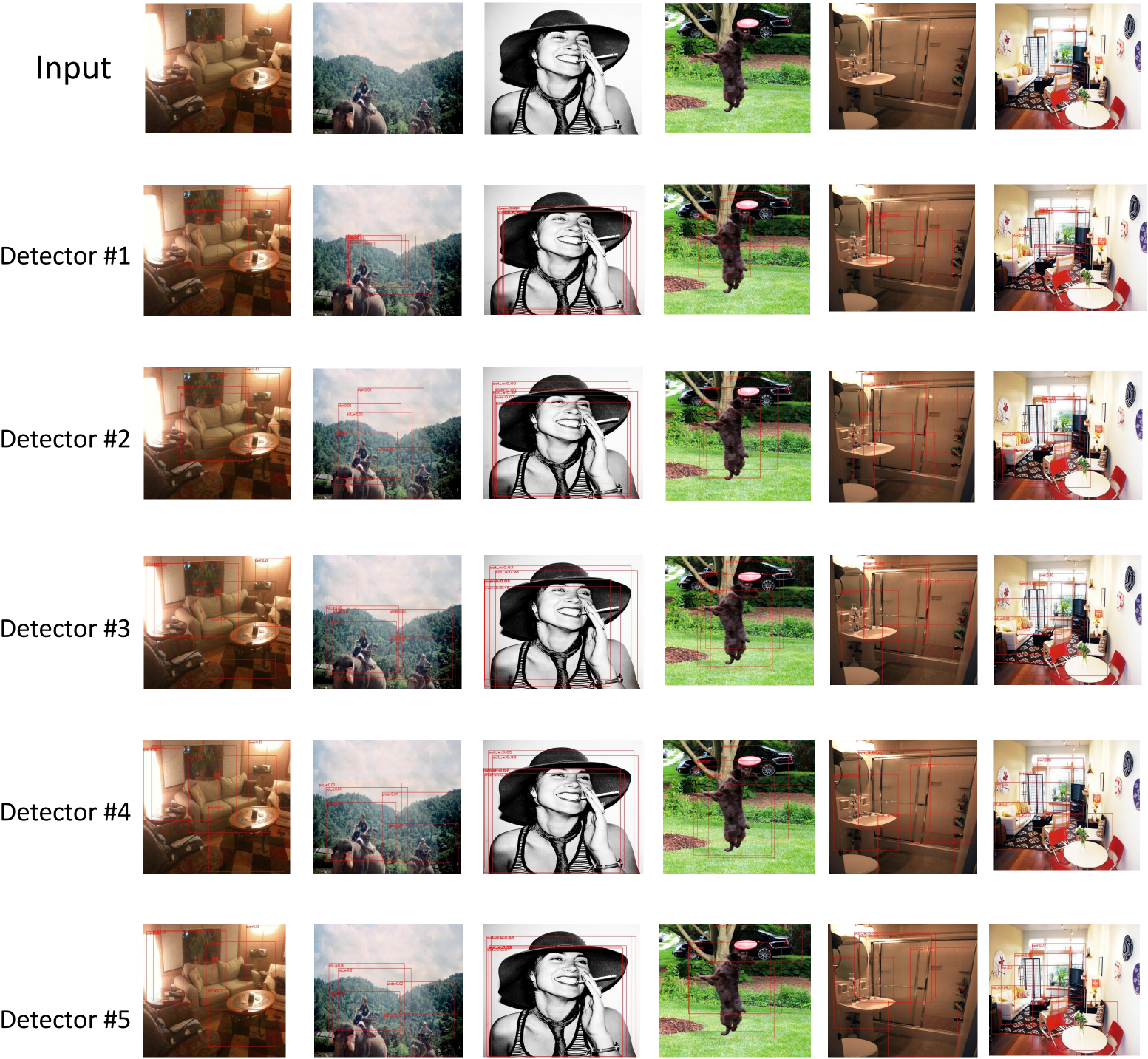}
    \vspace{-7pt}
    \caption{Detailed case study examples of the predicted results of each detector of \textit{VIGIL-8B} under the setting $M=N=5$. All images are resized to the same scale for better arrangement and visualization. The annotations in the images ``comp", ``over", ``under", ``sat-s", `sat-w", and ``over.sh" denote ``compression", ``overexposure-contrast-strengthen", ``underexposure-contrast-weaken", ``saturate-strength", ``saturate-weaken", and ``oversharpen", respectively.
} 
    \label{fig:case}
    
\end{figure*}

\begin{figure*}
    \centering
    \includegraphics[width=\linewidth]{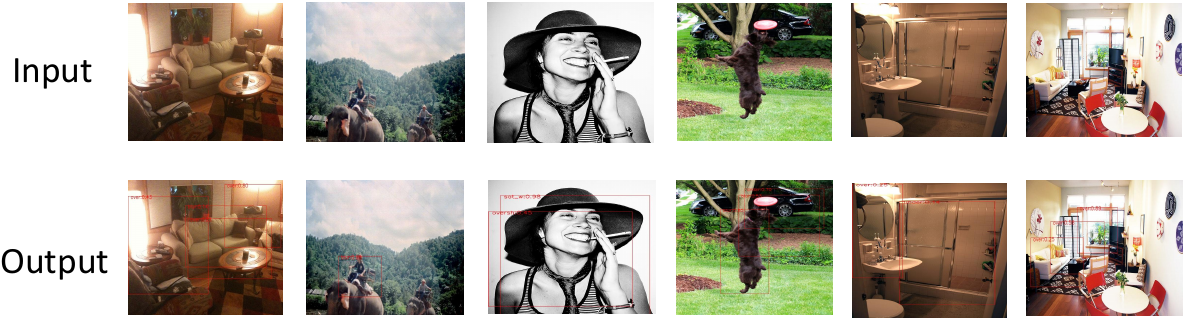}
    \vspace{-7pt}
    \caption{Examples of the final predicted results after post-processing of \textit{VIGIL-8B} under the setting $M=5,N=3$ (our setting in the main paper). All images are resized to the same scale for better arrangement and visualization.
} 
    \label{fig:case1}
    
\end{figure*}

In Fig.~\ref{fig:case}, we present the detection results for each detector of \textbf{VIGIL-8B} under the \(M = N = 5\) setting (the choice of \(N = 5\) instead of the main setting \(N = 3\) from the main paper is made to more clearly show the detection differences among the detectors). In Fig.~\ref{fig:case1}, we present the final output results of the model under the main setting (\(M = 5, N = 3\)) in the main paper.

\end{document}